\documentclass{article}

\usepackage{PRIMEarxiv}

\usepackage[utf8]{inputenc} 
\usepackage{graphicx}
\usepackage{subcaption}
\usepackage{float}
\usepackage[colorlinks=true,linkcolor=blue,citecolor=blue]{hyperref}

\usepackage{soul}
\usepackage{xcolor}

\usepackage{booktabs}
 \usepackage{graphics}
 \usepackage{tabularx}
\usepackage{makecell}
\usepackage[export]{adjustbox}
\usepackage{multirow}
\usepackage{enumitem}
\setlist{labelindent=12pt}
\usepackage{subcaption}
\usepackage{algorithm} 
\usepackage{algpseudocode}
\usepackage[labelfont=bf,textfont={bf}]{caption}
\usepackage{xcolor}
\usepackage{nccmath, mathtools}
\usepackage{svg}
\usepackage{amssymb} 
\usepackage{etoolbox}
\apptocmd\normalsize{%
 \abovedisplayskip=12pt
 \abovedisplayshortskip=0pt 
 \belowdisplayskip=12pt 
 \belowdisplayshortskip=7pt
}{}{}
\usepackage{mathtools}
\usepackage{physics}
\hypersetup{
    colorlinks=true,                
    breaklinks=true,                
    urlcolor= cyan,                
    linkcolor= red,                     
    bookmarksopen=false,
    filecolor=black,
    citecolor=blue,
    linkbordercolor=pink
}

\renewcommand{\mathbf}{\boldsymbol}

\newcommand{\mb}{\mathbf}
\newcommand{\mc}{\mathcal}

\newcount\Comments  
\usepackage{color}
\definecolor{darkgreen}{rgb}{0,0.5,0}
\definecolor{purple}{rgb}{1,0,1}
\newcommand{\kibitz}[2]{\ifnum\Comments=0\textcolor{#1}{#2}\fi}

\usepackage{tikz}

\title{RACER: Rational Artificial Intelligence Car-following-model Enhanced by Reality
\thanks{\textit{\underline{Citation}}: 
\textbf{Li, et al. RACER: Rational Artificial Intelligence Car-following-model Enhanced by Reality.}} 
}

\author{
  Tianyi Li \\
  Department of Civil Engineering \\
  Saint Louis University \\
  \texttt{tianyili.ai@gmail.com} \\
   \And
  Alexander Halatsis \\
  Department of Aerospace Engineering \\
  University of Minnesota \\
  500 Pillsbury Dr. SE, Minneapolis, MN 55455, USA.\\
  \texttt{Halat005@umn.edu} \\
  \AND
  Raphael Stern \\
  Department of Civil, Environmental, and Geo-Engineering \\
  University of Minnesota \\
  500 Pillsbury Dr. SE, Minneapolis, MN 55455, USA.\\
  \texttt{rstern@umn.edu} \\
}

\begin{document}
\maketitle

\begin{abstract}
This paper introduces RACER, the Rational Artificial Intelligence Car-following model Enhanced by Reality, a cutting-edge deep learning car-following model, that satisfies partial derivative constraints, designed to predict Adaptive Cruise Control (ACC) driving behavior while staying theoretically feasible. Unlike conventional models, RACER effectively integrates Rational Driving Constraints (RDCs), crucial tenets of actual driving, resulting in strikingly accurate and realistic predictions. Against established models like the Optimal Velocity Relative Velocity (OVRV), a car-following Neural Network (NN), and a car-following Physics-Informed Neural Network (PINN), RACER excels across key metrics, such as acceleration, velocity, and spacing. Notably, it displays a perfect adherence to the RDCs, registering zero violations, in stark contrast to other models. This study highlights the immense value of incorporating physical constraints within AI models, especially for augmenting safety measures in transportation. It also paves the way for future research to test these models against human driving data, with the potential to guide safer and more rational driving behavior. The versatility of the proposed model, including its potential to incorporate additional derivative constraints and broader architectural applications, enhances its appeal and broadens its impact within the scientific community.
\end{abstract}


\section{Introduction}
The revolutionary advances in vehicle automation have significant implications for the transportation sector. Numerous studies have delved into automated vehicle (AV) related topics such as enhanced traffic~\cite{tan1998demonstration, wang2022optimal, wang2021optimal} and speed harmonization~\cite{learn2017freeway}, among others. Although certain aspects of vehicle automation may augment traffic flow~\cite{stern2017dissipation, talebpour2016influence}, not all impacts are advantageous. For instance, existing research demonstrates that commercially available adaptive cruise control (ACC) vehicles may potentially decrease highway throughput~\cite{shang2021impacts}. These findings emphasize the significance of individual vehicles' driving characteristics in determining the aggregate behavior of the overall traffic flow. Consequently, accurately calibrated car-following models are an essential tool for analyzing the influence of ACC and AV dynamics on traffic flow and stability~\cite{talebpour2015influence, shang2021impacts, davis2004effect, davis2013effects, gunter2020are, shang2022novel}.

While many different approaches have been proposed to model the vehicle-level dynamics of automated and partially automated vehicles~\cite{talebpour2016influence,davis2004effect, milanes2014modeling}, the majority of these models adapt existing car-following models to capture the dynamics of automated driving systems or AVs.
While full automation in vehicles is likely to be a reality at some point in the future, current trends already showcase widespread usage of lower levels of autonomy such as driver-assist features like ACC. The introduction of new traffic dynamics from ACC vehicles is expected to impact traffic flow, even at low ACC market penetration rates~\cite{gunter2020are, shang2021impacts}. Furthermore, these dynamics would be contingent on the type of ACC vehicle and its driving patterns.

Since the early General Motors car following experiments in the 1950s~\cite{gazis1959car}, car-following (CF) behavioral modeling has been the focal point of the transportation community to decipher how human drivers react to their surrounding traffic environment. With the advent of deep learning, its application in car-following modeling has also seen significant traction~\cite{wang2017capturing, panwai2007neural,zhu2018human, mo2021physics, naing2022dynamic, li2023car}. Deep learning~\cite{goodfellow2016deep}, with its ability to learn hierarchical representations autonomously from raw data, has spurred a revolution in the way data is interpreted across diverse fields such as image recognition and natural language processing~\cite{vaswani2017attention,deng2009imagenet, schmidhuber2015deep}. But, it also brings to the table certain challenges like overfitting and lack of interpretability, leading to the development of models like physics-informed neural networks (PINN)~\cite{raissi2019physics}.

The CF modeling methods landscape is broadly divisible into three distinct groups~\cite{mo2021physics}: physics-based models, data-driven models, and the emerging class of physics-guided AI car-following models. Physics-based models~\cite{gunter2020are, shang2022novel, talebpour2016influence, treiber2000congested, jiang2001full} rely on predefined mathematical functions with a limited number of parameters. These models simplify interactions between drivers and other vehicles, by using a simplified representation of human cognitive processes and mechanical dynamics. Data-driven models, on the other hand, have the unique capability to learn hierarchical representations directly from raw data, eliminating the need for explicit feature extraction or model formulation~\cite{wang2017capturing, zhu2018human, panwai2007neural}.

Recent years have witnessed the emergence of a combination of physics and data-driven models, offering a promising pathway to develop robust and interpretable CF models. These models strive to amalgamate domain-specific knowledge with deep learning prowess, thereby harnessing the strengths of both to create models adept at handling complex driving scenarios while adhering to real-world physics principles~\cite{raissi2019physics, mo2021physics, naing2022dynamic, ma2023physics, ruan2022causal}. However, these models' significant limitation is their inability to comprehend a rational driver's actions in full. Both learning-based models and PINN works have overlooked Rational Driving Constraints (RDCs)~\cite{wilson2011car}, a set of conditions on car following behavior that are generally accepted as necessary for sensible or rational driving behavior, and traditional physics-based models such as OVRV can only represent a simplified version of reality, excluding many complexities in driving behavior~\cite{wang2017capturing, mo2021physics, naing2022dynamic, zhu2018human}.

Our approach addresses these limitations by incorporating domain-specific knowledge and physical derivative constraints into the neural network architecture. The objective is to endow the model with a comprehensive understanding of rational driving behavior while still leveraging the advantages of data-driven learning. Through this integration, our aspiration is to create a more robust and interpretable car-following model that surpasses the capabilities of existing approaches, laying the groundwork for safer and more reliable autonomous driving systems.

The key contributions of this study are threefold: 1) the introduction of a novel deep learning methodology that integrates the RDCs in car-following modeling, providing a more flexible and efficient car-following modeling paradigm while addressing existing models' limitations; 2) the demonstration of our model's superior performance when compared with existing physics-based, data-driven, and PINN models; 3) a comprehensive analysis showing that our proposed model satisfies the RDCs, a feat yet to be achieved by other machine learning models.

The remainder of this article unfolds as follows: In Section~\ref{sec:lit}, we introduce relevant literature at the intersection of car-following modeling and deep learning. We then explore various types of car-following models and elaborate on our proposed methodology and model in Section~\ref{sec:model}. Section~\ref{sec:data} presents the experimental data utilized in this study followed by the experiments in Section~\ref{sec:experiments}. Subsequently, Section~\ref{sec:analysis} provides a comprehensive discussion of our numerical experiment results and their implications. Finally, we conclude the article in Section~\ref{sec:con}, where we discuss the study's limitations and propose potential avenues for future research. To provide a clear understanding of the research problem and methodology, we present a comprehensive list of notations used throughout this paper in Table~\ref{tab:nomenclature}.

\begin{table}[htbp]
  \centering
  \caption{Nomenclature and acronyms used throughout the paper.}
  \label{tab:nomenclature}
  \footnotesize
  \setlength{\tabcolsep}{4pt}
  \renewcommand{\arraystretch}{1.15}
  \begin{tabular}{@{}llc@{}}
    \toprule
    \textbf{Symbol} & \textbf{Description} & \textbf{Unit} \\
    \midrule
    \multicolumn{3}{@{}l}{\textbf{Variables and Parameters}} \\ 
    \midrule
    $a_t,\;\hat a_t$            & True / predicted longitudinal acceleration              & m\,s$^{-2}$ \\
    $a_{\text{NN}},\,a_{\text{Phy}}$ & Acceleration predicted by NN / by physical model           & m\,s$^{-2}$ \\
    $f_{\theta}(\cdot)$         & Car-following mapping parameterized by $\theta$         & – \\
    $k_1,k_2$                   & OVRV gain parameters                                   & – \\
    $N$                         & Number of data points in a batch                        & – \\
    $s_t$                       & Space gap between follower and leader                   & m \\
    $\hat s(t)$                 & Measured inter-vehicle spacing                          & m \\
    $v_t$                       & Speed of follower vehicle                               & m\,s$^{-1}$ \\
    $\hat v_\ell(t)$, $\hat v_f(t)$ & Measured lead / following vehicle speed             & m\,s$^{-1}$ \\
    $X_{\text{seq}}, X_{\text{phy}}$ & Sequence input data / physical state input data    & – \\
    $\lambda_{1,2,3}$           & Weights on RDC penalties in loss function              & – \\
    $\tau, \eta$                & OVRV time constant / jam distance parameters            & s, m \\
    $\theta$                    & CF model parameters                                        & – \\
    $\mathcal{O},\;\mathcal{A}$ & Observation space / action space                        & – \\
    \midrule
    \multicolumn{3}{@{}l}{\textbf{Acronyms}} \\ 
    \midrule
    ACC   & Adaptive Cruise Control                                                   & – \\
    AV    & Automated Vehicle                                                         & – \\
    CF    & Car-Following                                                             & – \\
    LSTM  & Long Short-Term Memory network                                            & – \\
    MSE   & Mean Squared Error                                                        & – \\
    OVRV  & Optimal Velocity–Relative Velocity model                                  & – \\
    PINN  & Physics-Informed Neural Network                                           & – \\
    RDC   & Rational Driving Constraint                                               & – \\
    RMSE  & Root Mean Squared Error                                                   & – \\
    \bottomrule
  \end{tabular}
\end{table}

\section{Review of Relevant Literature}\label{sec:lit}
In this section, we first delve into the literature encompassing physically-based car-following models, examining their evolution and impact. Subsequently, we pivot our attention towards the significant body of work wherein deep learning has been employed to model driving behavior, and where it has been ingeniously amalgamated with traditional models, to address their individual limitations and optimize performance.

\subsection{The Modeling of Driving Behavior}

The quest to model individual vehicle dynamics has been an intriguing focus for researchers since the mid-20th century~\cite{gazis1959car}. The pivotal premise energizing this research area is that the conduct of one vehicle (dubbed as the ``following" vehicle in this context) is significantly influenced by the actions of the vehicle preceding it (the ``lead" vehicle).

In general, the following vehicle's acceleration at a specific time, $\ddot{x}(t)$, is expressed as a second-order ordinary differential equation, hinging on a selection of parameters: inter-vehicle spacing $s(t)$, the following vehicle speed $v(t) = \dot{x}(t)$, and the relative speed or inter-vehicle velocity $\dot{s}(t) = \dot{x}_\ell(t) - v(t)$~\cite{brackstone1999car}. In this expression, $\dot{x}_\ell(t) = {v}_\ell(t)$ signifies the lead vehicle's speed, whereas $x$ represents the distance or position from a predetermined starting point.

These well-known car-following models (CFMs) have evolved significantly over time~\cite{helbing2001traffic, Treiber_2013_book}. Classic models, such as the Gazis-Herman-Rothery (GHR) model~\cite{gazis1961nonlinear} and Optimal Velocity (OV) model~\cite{bando1995dynamical}, offer fairly simple mathematical formulations for the acceleration function $f$. In contrast, modern models like the Intelligent Driver Model (IDM)~\cite{treiber2000congested} and Full Velocity Difference Model (FVDM)~\cite{jiang2001full} incorporate additional parameters to accommodate subtler driving behaviors. These models establish deterministic relationships that accurately capture the impact of a lead vehicle's actions on a following vehicle, integrating aspects such as inter-vehicle distance, the speed of the following vehicle, and the relative speed between both vehicles.

Nevertheless, these models often face criticism for their oversimplification and inability to adapt to unique driving behavior~\cite{punzo2011influence}. Given these models are founded on predetermined assumptions and parameters, they may inadequately capture the variability and intricacy of real-world driving behaviors, especially in dynamic traffic scenarios~\cite{kesting2008variable}.

In response to these limitations, contemporary research is gravitating towards the integration of machine learning techniques for modeling driving behavior. These machine learning models, capable of learning directly from data, hold the potential to adapt to the patterns and variability inherent in individual drivers' behavior. This shift has spurred the creation of innovative models such as the PINN based car-following models.

\subsection{The Realm of Deep Learning and Driving Behavior}
In the realm of driving behavior modeling, deep learning algorithms have increasingly been utilized to extract intricate car-following behaviors directly from trajectory data~\cite{hoel2018automated}. Furthermore, deep reinforcement learning has proven effective, particularly for large-scale simulations of naturalistic driving environments involving multiple objects~\cite{yan2023learning, feng2023dense}. Certain studies, such as the neural network-based controller for ACC vehicles, have demonstrated superior performance in minimizing the discrepancy between real trajectories and model predictions~\cite{mahadika2020neural}. However, they bring along a unique set of challenges. Deep learning models typically require extensive data and are vulnerable to overfitting, particularly when used in control settings, leading to less versatile control models~\cite{mahadika2020neural}. Furthermore, the output generated by these models often lacks interpretability, leading to their perception as ``black boxes"~\cite{samek2017explainable}. In control environments, these ``black boxes" play a crucial role in the car's unpredictable behavior, where the addition of more layers may lead to less safe behavior~\cite{cherian2012neural}.

To circumvent these challenges, hybrid models such as PINN-CFM have been conceived~\cite{mo2021physics,naing2022dynamic, ma2023physics}. By infusing physics-based components into the deep learning architecture, these models boost their versatility and interpretability. This integration cleverly combines the strengths of both deep learning and physics-based modeling, while simultaneously curbing some of their inherent limitations. Cherian's study revealed that increasing the neuron count augments driving comfort but at the cost of compromised safety and elevated speed deviation. This underscores the criticality of integrating physical constraints into the neural network training process. It ensures that safe and comfortable driving predictions are based on the car's actual behavior, rather than solely depending on the training data~\cite{cherian2012neural}.

Despite this progress, contemporary models still harbor certain limitations. Firstly, they typically rely on a specific physical car-following model, such as the IDM, in their loss functions, striving to find the best fit for physical car-following models~\cite{mo2021physics, naing2022dynamic, ma2023physics}. Secondly, they fall short of fully accounting for the variation among drivers. Moreover, these models may fail to adhere to RDCs~\cite{wilson2011car}.

In response to these limitations, this study introduces a novel deep-learning model rooted exclusively in the partial derivative constraints of the RDCs. This model presents a more flexible and efficient constraint compared to those used in PINN-CF models. Consequently, we introduce a rational neural network-based car-following model, potentially transcending the capabilities of existing models.

\section{Modeling Car-following Behavior}\label{sec:model}

In this section, we propose a rational neural network-based car-following model designed to meet the RDCs while being capable of adapting to varying driving conditions and unique driver behaviors. We will detail the model's architecture, data processing, training process, and alignment with the RDC. Before delving into the proposed model, we first present an overview of the car-following model and three distinct types of modeling techniques. The classification of the driving policy mapping models includes engineering, deep learning (DL)-based, and physics-guided AI car-following models.

A car-following model is a mathematical mapping parameterized by $\mb \theta$, represented as $\mb f_{\theta}(\cdot|\theta)$, that maps state variables (observations of the traffic environment) $o \in \mc {O}$ to actions $a \in \mc {A}$ (longitudinal accelerations):

\begin{equation}
f_{\theta} : o \rightarrow a
\end{equation}

While a vast body of literature employs various variables as inputs, this study focuses on three, namely the spacing $ {s}(t)$, relative speed $\Delta {v}(t)$, and the subject vehicle's velocity $ {v}(t)$ at the current time step. The state variable $\mb o$ constitutes a 3-dimensional state vector: $o = ({s}(t),\Delta{v}(t),{v}(t))$. The acceleration $a(t)$ of a following vehicle at time $t$ is depicted as a second-order ordinary differential equation, a function of the inter-vehicle spacing $s(t)$, the following vehicle speed $v(t) = \dot{x}(t)$, and the relative speed or inter-vehicle velocity $\dot{s}(t)$. Here, $x$ symbolizes the position or distance from an arbitrary starting point, and the derivative relationships are:

\begin{equation}\label{eq:der_1}
v(t) = \dot{x}(t)
\end{equation}

\begin{equation}\label{eq:der_2}
\Delta{v}(t) = \dot{s}(t)
\end{equation}

\begin{equation}\label{eq:der_3}
a(t) = \dot{v}(t) = \ddot{x}(t)
\end{equation}

This CFM can then be formulated as:
\begin{equation}\label{eq:dynamics}
\ddot{x}(t) = f_{\theta}(s(t), \Delta{v}(t), {v}(t))
\end{equation}

\subsection{Engineering Car-following Model}
The input of the Physics-based Car-following Model (Phy-CFM) or Engineering Car-following Model includes all possible signals $s(t)$ at time $t$ from neighboring vehicles in the traffic environment, and its output is the acceleration $a(t +\Delta t)$ at the subsequent time step. The Phy-CFM model aims to find an optimal set of parameters $\theta^*$ that best fits real-world data to minimize the error between a simulated trajectory for a following vehicle and the observed vehicle trajectory as measured in a dataset. One approach that has proven to be successful at finding best-fit model parameter values is to minimize the mean square error (MSE) in inter-vehicle spacing between the simulated car following trajectory and the inter-vehicle spacing in the experimental data~\cite{li2021classification, li2023car}. The procedure can be written as:

\begin{equation}\label{eq:phy_CFM}
\begin{array}{l}
\underset{\theta}{{\text{minimize}}}: \sum_{t=0}^N (\hat{a}(t) - {a}_{Phy}(t))^2  \\
\text{s.t.} \quad a(t + \Delta t) =  f_{\theta}(s(t), \Delta{v}(t), \dot{v}(t)), \quad t = 0, \Delta t, \ldots, N
\end{array}
\end{equation}

In this context, $\hat{a}(t)$ denotes the actual acceleration at time $t$, ${a}_{Phy}(t)$ represents the acceleration predicted by the physical model at time $t$, $f_{\theta}(\cdot)$ is the physical model of the prediction function parameterized by $\theta$, $\Delta t$ is the time step, and $N$ is the size of the provided observed data. In most scenarios, the optimization problem in Equation~\eqref{eq:phy_CFM} can be efficiently solved via numerical optimization.

\subsection{Deep Learning based Car-following Model}
While physics-based mappings are typically represented by mathematical formulas, deep learning-based mappings are often embodied through deep neural networks. The Artificial Neural Network (ANN) is a commonly used function approximator for car-following behavior. It considers states as inputs and generates the corresponding accelerations as outputs. Within each layer, every node linearly combines the outputs from the preceding layer and applies a non-linear activation function $\sigma$, passing it to the subsequent layer.

The mathematical representation of an ANN is as follows:

\begin{equation}
a_{NN} = \sigma\left( I \right)= \sigma\left( \mathbf{W}_L \left( \dots \left( \mathbf{W}_1 \mathbf{o} + \mathbf{b}_1 \right) \dots + \mathbf{b}_{L-1} \right) + \mathbf{b}_L \right)
\end{equation}
Where $\sigma(\cdot)$ denotes the activation function, $\mathbf{W}_i$ represents the weight matrix for the $i$-th layer, $\mathbf{b}_i$ is the bias for the $i$-th layer, and $\mathbf{o}$ refers to the input vector to the network. This equation illustrates the network's transformation, i.e., the output of the network given the input state $\mathbf{o}$.

The application of machine learning techniques in modeling, classifying, and simulating car-following behaviors has garnered significant attention in recent years~\cite{wang2017capturing, li2022detecting, li2022robustness, li2023car, li2024customizable}. These advanced systems have demonstrated superior performance over traditional physical models. However, they often grapple with challenges of interpretability and may exhibit limitations in handling scenarios that diverge from their training datasets.

\subsection{Physics-guided AI Car-following Model}

{A major shortcoming of learning-based models is their inability to consistently yield physically plausible outputs. This issue could lead to unsafe and irrational driving behaviors resulting in vehicle collisions in microscopic traffic simulations~\cite{willard2020integrating}. However, the recent development of PINNs offers a promising path toward a data-driven modeling paradigm that leverages the strengths of both physical and deep learning models~\cite{raissi2019physics}. A step in this direction is the car-following model proposed by Mo et al., which combines physical models with learning-based techniques~\cite{mo2021physics, naing2022dynamic}. The key distinction between PINN-CFM and purely neural network-based CFM lies in the loss function, modified to include a weighted sum of two components: the data discrepancy loss $MSE_{NN}$ calculated by the neural network, and the physics discrepancy estimated using the physical car-following model $MSE_{Phy}$.}

\begin{equation}
\begin{aligned}
L_{\theta} &= \alpha \cdot MSE_{NN} + (1 - \alpha) \cdot MSE_{Phy} \\
&= \alpha \cdot \left( \frac{1}{N} \sum_{i=1}^{N} \left\| \hat{a}{(i)} - {a}_{NN}^{(i)} \right\|^2 \right) 
\\ 
&+ (1 - \alpha) \cdot \left( \frac{1}{N} \sum_{i=1}^{N} \left\| {a}_{NN}^{(i)} - {a}_{Phy}^{(i)} \right\|^2 \right)
\label{CFM:PINN}
\end{aligned}
\end{equation}

Here, $L_{\theta}$ denotes the loss function, $MSE_{NN}$ and $MSE_{Phy}$ represent the mean square errors for data discrepancy and physics discrepancy, respectively, $\alpha$ is the trade-off coefficient, and $N$ represents the size of observed data. The actual acceleration and its estimate for the observed state $i$ are denoted as $a{(i)}$. Finally, ${a}_{NN}^{(i)}$ and ${a}_{Phy}^{(i)}$ express the estimated accelerations from the neural network model and engineering models, respectively.

Nevertheless, the existing models fall short in accounting for RDCs. The final outputs of these models are solely determined by learning-based mechanisms, and thus, cannot guarantee the rationality of the learned model. As a result, scenarios with potential collisions and unsafe driving behaviors could still occur, signaling a critical area for further research and improvement~\cite{mo2021physics, ma2023physics}.

\subsection{Deep Learning-Based Car-Following Model Integration of Rational Driving Constraints}

Previous car-following models relying on machine learning, as referenced in \cite{panwai2007neural, huang2018car, wang2017capturing}, have exhibited high performance in certain metrics, notably the MSE in predicted accelerations. Despite their capabilities, these models, including those developed on PINN~\cite{mo2021physics, naing2022dynamic, ma2023physics} and reinforcement learning \cite{zhu2018human}, often neglect the integration of RDCs. This section aims to fill this gap by outlining a deep learning-based car-following model that efficiently integrates RDC. As far as we know, this is an innovative addition to the field

Rational Driving Constraints are critical safety constraints for car-following models that are used to control autonomous vehicles. They represent natural behavioral laws that all rational drivers would obey to maintain safe driving~\cite{wilson2011car, stern2017dissipation}. Specifically, these constraints state that a rational driver would always try to:

\begin{itemize}
\item \textbf{Speed Constraint}: Ensure that the derivative of acceleration with respect to speed is non-positive, indicating that the vehicle does not accelerate at higher speeds. This is mathematically represented as:

\begin{equation}
\frac{da}{dv} \leq 0
\label{RDC_1}
\end{equation}

\item \textbf{Spacing Constraint}: Ensure that the derivative of acceleration with respect to spacing is non-negative, indicating that the vehicle does not decelerate when the spacing increases. This is mathematically represented as:

\begin{equation}
\frac{da}{ds} \geq 0
\label{RDC_2}
\end{equation}

\item \textbf{Relative Speed Constraint}: Ensure that the derivative of acceleration with respect to relative speed is non-negative, indicating that the vehicle does not decelerate when the relative speed increases. This is mathematically represented as:

\begin{equation}
\frac{da}{dr} \geq 0
\label{RDC_3}
\end{equation}
\end{itemize}

Implementing these constraints in learning-based car-following models is essential as they warrant that the model's predictions align with basic safe and rational driving principles. The proposed solution in the provided code introduces a novel approach to enforce RDCs by incorporating them into the loss function of the model. This effectively guides the model during training in learning predictions that satisfy RDCs.

Our proposed model termed the ``RACER: Rational Artificial Intelligence Car-following-model Enhanced by Reality," is a synergistic combination of two integral components. The first is a neural network component that caters to the sequential patterns within the data. The second is the incorporation of RDCs within the loss function. This ensures the model's predictions align with the real-world physics principles of vehicle operation. This is particularly critical for automated vehicles, which require adherence to rational and safe driving principles.

The RDCs are integrated into the model by formulating a custom loss function that ensures adherence to real-world physics principles. Specifically, the RDCs constraints are enforced by adding penalty terms into the loss function that measure the degree of violation of each constraint. Specifically, we define our loss as a combination of MSE and an RDCs-informed constraint term, defined as follows:

\begin{equation}
\begin{aligned}
Loss_{RDCs} &= MSE_{NN} + \lambda \cdot MSE_{RDCs}
\\
&= MSE(a_{pred}, a_{true}) + \lambda_1 \cdot RDC_{speed} 
\\
&+ \lambda_2 \cdot RDC_{spacing} + \lambda_3 \cdot RDC_{\text{relative speed}}
\\
&= \frac{1}{N} \sum_{i=1}^{N} \left(a_{\text{true}}^{(i)} - a_{\text{pred}}^{(i)}\right)^2 
\\
&+ \lambda_1 \cdot \left( \frac{1}{N} \sum_{i=1}^{N} \text{ReLU}\left(\frac{\partial a_{\text{pred}}^{(i)}}{\partial v^{(i)}}\right) \right)
\\
&+ \lambda_2 \cdot \left( \frac{1}{N} \sum_{i=1}^{N} \text{ReLU}\left(-\frac{\partial a_{\text{pred}}^{(i)}}{\partial s^{(i)}}\right) \right)
\\
&+ \lambda_3 \cdot \left( \frac{1}{N} \sum_{i=1}^{N} \text{ReLU}\left(-\frac{\partial a_{\text{pred}}^{(i)}}{\partial r^{(i)}}\right) \right)
\end{aligned}
\end{equation}

In this equation, $a_{\text{true}}^{(i)}$ and $a_{\text{pred}}^{(i)}$ represent the actual and predicted acceleration at the $i^{th}$ time step, and $\lambda$ is the trade-off coefficient respectively. The terms $\frac{\partial a_{\text{pred}}^{(i)}}{\partial v^{(i)}}$, $\frac{\partial a_{\text{pred}}^{(i)}}{\partial s^{(i)}}$, and $\frac{\partial a_{\text{pred}}^{(i)}}{\partial r^{(i)}}$ represent the derivatives of the predicted acceleration concerning velocity, spacing, and relative velocity at the $i^{th}$ time step, respectively. Here, $v^{(i)}$, $s^{(i)}$, and $r^{(i)}$ correspond to velocity, spacing, and relative velocity at the $i^{th}$ time step, respectively, while $N$ signifies the total number of data points in a batch.

This loss function is crafted to minimize the MSE between the predicted and true accelerations while also respecting the RDCs, mediated by the hyperparameters $\lambda_1$, $\lambda_2$, and $\lambda_3$. The constraints are modeled as $RDC_{speed}$, $RDC_{spacing}$, and $RDC_{\text{relative speed}}$, which are computed as the derivatives of the model's output corresponding to the driving parameters. By applying the ReLU function, these quantities are ensured to be non-positive.

In our car-following model, let's denote the complex function that maps inputs like speed ($v$), spacing ($s$), and relative speed ($r$) to the predicted acceleration ($a_{\text{pred}}$) as $f(x)$. The function $f(x)$ includes elementary operations, denoted as $g$, that could represent various transformations and activations in a deep learning model. These operations are the fundamental building blocks that we leverage when performing automatic differentiation, which is a process that follows the chain rule for differentiation in an efficient manner. Specifically, for an arbitrary function $f(x)$ that is composed of these elementary operations $g$, automatic differentiation is expressed as follows:

\begin{equation}
\frac{df}{dx} = \frac{df}{dg} \cdot \frac{dg}{dx}
\end{equation}

In this process, the chain rule is recursively applied, enabling us to compute the derivatives of $a_{\text{pred}}$ with respect to $v$, $s$, and $r$ that form the basis for the computation of RDCs. Once these derivatives are obtained, we use the Rectified Linear Unit (ReLU) function~\cite{agarap2018deep}, $\text{ReLU}(x) = max(0, x)$ to retain only those portions of the gradients that violate the rational driving constraints, as the RDCs are presented in Equations~\eqref{RDC_1}-\eqref{RDC_3}, with the ReLU function serving to retain only those gradients violating the RDCs, i.e., the positive velocity gradients and the negative gradients of spacing and relative speed. This innovative application of automatic differentiation and the ReLU function in the computation of RDCs allows the integration of these constraints into the loss function, thereby guiding the model's predictions to align more closely with the principles of vehicular movement.

Using the calculated gradients, the ReLU function is applied as follows:

\begin{equation}
RDC_{speed} = \text{ReLU}\left(\frac{\partial a_{\text{pred}}}{\partial v}\right) = \max\left(0, \frac{\partial a_{\text{pred}}}{\partial v}\right)
\end{equation}

\begin{equation}
RDC_{spacing} = \text{ReLU}\left(-\frac{\partial a_{\text{pred}}}{\partial s}\right) = \max\left(0, -\frac{\partial a_{\text{pred}}}{\partial s}\right)
\end{equation}

\begin{equation}
RDC_{\text{relative speed}} = \text{ReLU}\left(-\frac{\partial a_{\text{pred}}}{\partial r}\right) = \max\left(0, -\frac{\partial a_{\text{pred}}}{\partial r}\right)
\end{equation}

These operations enforce the RDCs, retaining only those portions of the gradients that violate the constraints while nullifying the rest.
By integrating these RDCs into the loss function, the model's predictions are guided to align with the principles of vehicular movement, producing more realistic results. This novel integration of domain-specific knowledge is a significant step towards developing more robust and reliable car-following models.

In our proposed model, two distinct inputs, denoted as $X_{\text{seq}}$ and $X_{\text{phy}}$, are utilized. The $X_{\text{seq}}$ input includes time-dependent variables (speed, spacing, and relative speed) and is processed through several layers of a neural network architecture that can handle sequence data to manage temporal dependencies in driving behavior. This processing allows us to capture the intricate temporal dynamics that occur in vehicle-following scenarios. On the other hand, $X_{\text{phy}}$, containing speed ($v$), spacing ($s$), and relative speed ($r$) at a particular timestep, is used to enforce the RDCs through the calculation of gradients. To apply the RDCs in our loss function, we compute the derivatives of $a_{\text{pred}}$ with respect to $v$, $s$, and $r$. These derivatives represent the gradients of the prediction contingent on each variable in $X_{\text{phy}}$. 

\begin{figure}[htbp]
    \centering
    \includegraphics[width=0.49\textwidth]{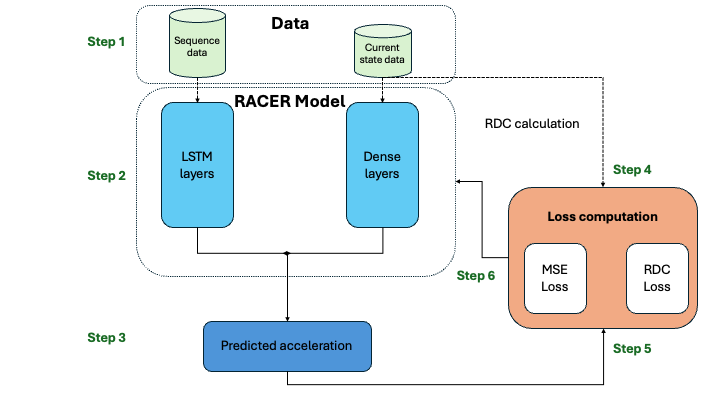}
    \caption{Architecture of the RACER model showing end-to-end dataflow: 
    (1) input encoding of sequence data ($X_{\text{seq}}$) and kinematic states ($X_{\text{phy}}$); 
    (2) neural network processing; 
    (3) acceleration prediction ($a_{\text{pred}}$); 
    (4) derivative computation for RDC calculation; 
    (5) composite loss formation; 
    (6) backpropagation. 
    Solid arrows indicate forward flow; dashed arrows show gradient computations.}
    \label{fig:racer-architecture}
\end{figure}

The overall architecture of the RACER model is illustrated in Fig.~\ref{fig:racer-architecture}. As shown, the model processes two distinct data inputs: sequential data ($X_{\text{seq}}$) and physical state data ($X_{\text{phy}}$). The process begins with \textbf{(1) input encoding} of this data, which then flows into \textbf{(2) the neural network processing} core. This pathway generates \textbf{(3) an acceleration prediction ($a_{\text{pred}}$)}. Simultaneously, this prediction, along with current state variables, is used for \textbf{(4) derivative computation}, enabling the calculation of both the MSE loss and the RDC loss. These are combined into \textbf{(5) a composite loss}, which is then used for \textbf{(6) backpropagation} to update the model parameters. This end-to-end process, guided by both forward data flow (solid arrows) and gradient/derivative information (dashed arrows), ensures both predictive accuracy and rational driving constraint compliance.

\begin{algorithm}
\caption{Enforcing Rational Driving Constraints}
\label{RDC-alg}
\begin{algorithmic}[1]
\State \textbf{Input:} Sequence of vehicle states $X_{\text{seq}}$, Physical vehicle states $X_{\text{phy}}$
\State \textbf{Output:} Predicted acceleration $a_{\text{pred}}$, Loss $\mathcal{L}$
\State $Z_{\text{seq}} \leftarrow NN_{\text{seq}}(X_{\text{seq}})$  \Comment{LSTM encoding}
\State $Z_{\text{seq}}' \leftarrow \text{process}(Z_{\text{seq}})$ \Comment{Further processing}
\State $Z_{\text{phy}}' \leftarrow \text{process}(X_{\text{phy}})$ \Comment{Physical state processing}
\State $a_{\text{pred}} \leftarrow \text{combine}(Z_{\text{seq}}', Z_{\text{phy}}')$ \Comment{Combine representations}
\State Calculate gradients $\frac{\partial a_{\text{pred}}}{\partial v}$, $\frac{\partial a_{\text{pred}}}{\partial s}$, $\frac{\partial a_{\text{pred}}}{\partial r}$ 
\State $RDC_{\text{speed}} \leftarrow \text{ReLU}\left(\frac{\partial a_{\text{pred}}}{\partial v}\right)$ \Comment{Speed constraint}
\State $RDC_{\text{spacing}} \leftarrow \text{ReLU}\left(-\frac{\partial a_{\text{pred}}}{\partial s}\right)$ \Comment{Spacing constraint}
\State $RDC_{\text{rel\_speed}} \leftarrow \text{ReLU}\left(-\frac{\partial a_{\text{pred}}}{\partial r}\right)$ \Comment{Relative speed constraint}
\State $\mathcal{L} \leftarrow \text{MSE}(a_{\text{pred}}, a_{\text{true}}) + \lambda_1 \cdot RDC_{\text{speed}} + \lambda_2 \cdot RDC_{\text{spacing}} + \lambda_3 \cdot RDC_{\text{rel\_speed}}$ \Comment{Combined loss}
\State \textbf{return} $a_{\text{pred}}, \mathcal{L}$
\end{algorithmic}
\end{algorithm}

The details of the algorithm are presented in Algorithm~\ref{RDC-alg} and Fig.~\ref{fig:racer-architecture}. The algorithm operates through several key stages: $NN_{\text{seq}}$ represents the sequence handling neural network (line 3), $\text{process}(\cdot)$ in lines 4-5 signifies additional processing steps that may vary depending on the exact network architecture used, and $\text{combine}(\cdot)$ in line 6 refers to the operation that combines the processed sequence and physical inputs. The core innovation lies in using two distinct inputs: $X_{\text{seq}}$ (temporal sequence data) and $X_{\text{phy}}$ (current kinematic state data typically used in physical models). Lines 7-10 implement the RDC enforcement by computing partial derivatives and applying ReLU functions to penalize constraint violations. The final loss function (line 11) combines the MSE loss for prediction accuracy with the RDC loss for constraint compliance. This integration of domain-specific knowledge and sequence data modeling represents a crucial step toward developing more reliable car-following models.

\section{Data Description}\label{sec:data}

Our analysis employs a principal dataset derived from a sequence of car-following experiments conducted by Gunter et al.~\cite{gunter2020are}. This dataset is amassed using a variety of commercially available vehicles equipped with ACC systems. Each of these ACC-activated vehicles adheres to a uniform testing procedure, where a leading vehicle traverses at a pre-established speed sequence for a set duration at each pace. The ACC vehicle, while trailing the lead vehicle, has its ACC active throughout the experiment. To accurately emulate car-following behavior, the acceleration of the trailing vehicle is determined by evaluating its speed change over a time interval, corresponding to some time steps as the data is captured at a rate of 10 Hz (0.1 s intervals). The whole dataset includes about 10 minutes of vehicle driving trajectory, resulting in more than 6000 rows of data.

\begin{figure}
\centering
\subfloat[Lead vehicle and follow vehicle speed (min setting).]{\includegraphics[scale = 0.14]{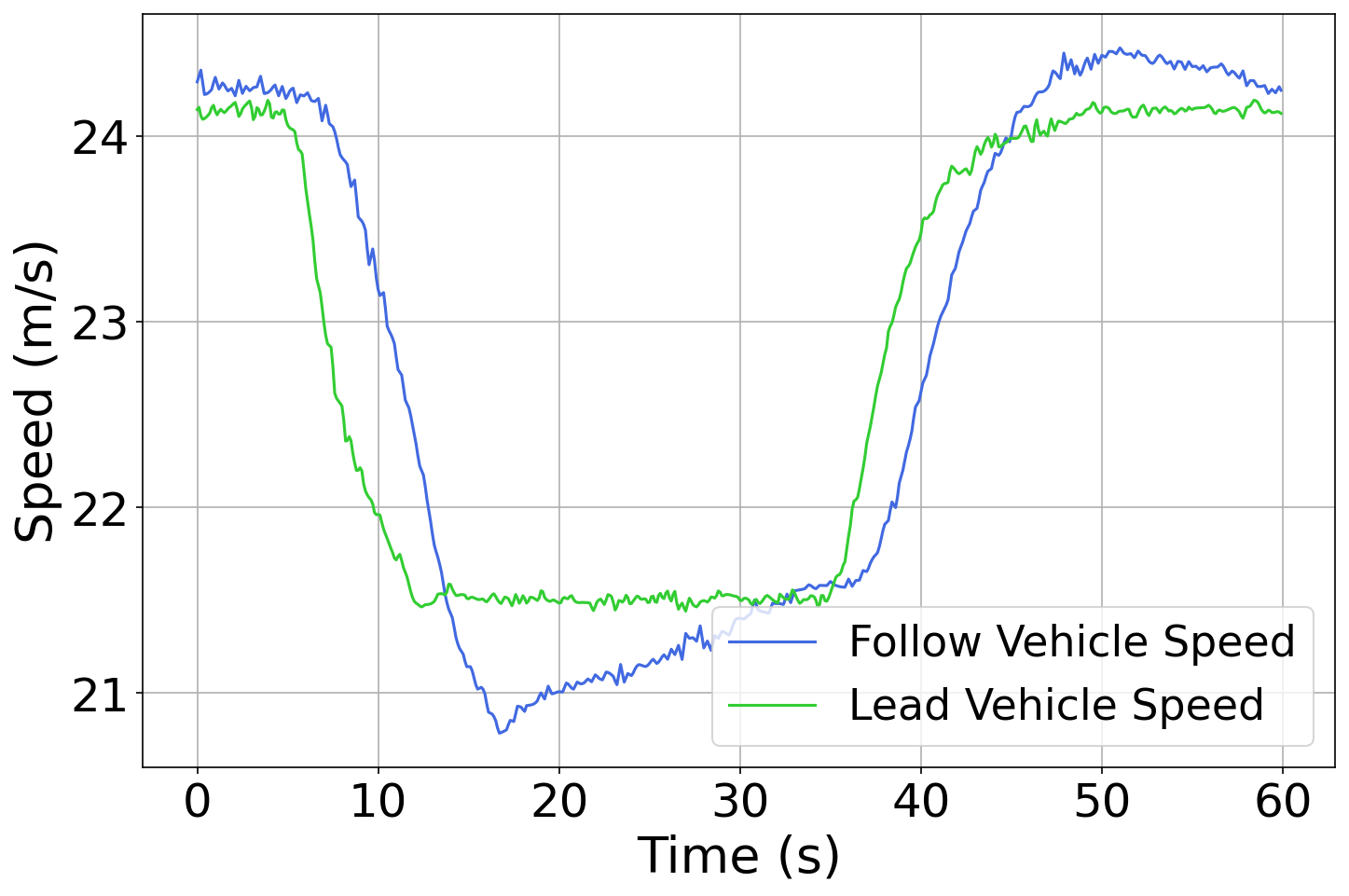}%
\label{fig_first_case}}
\hspace*{\fill} 
\subfloat[Inter-vehicle spacing (min setting).]{\includegraphics[scale = 0.14]{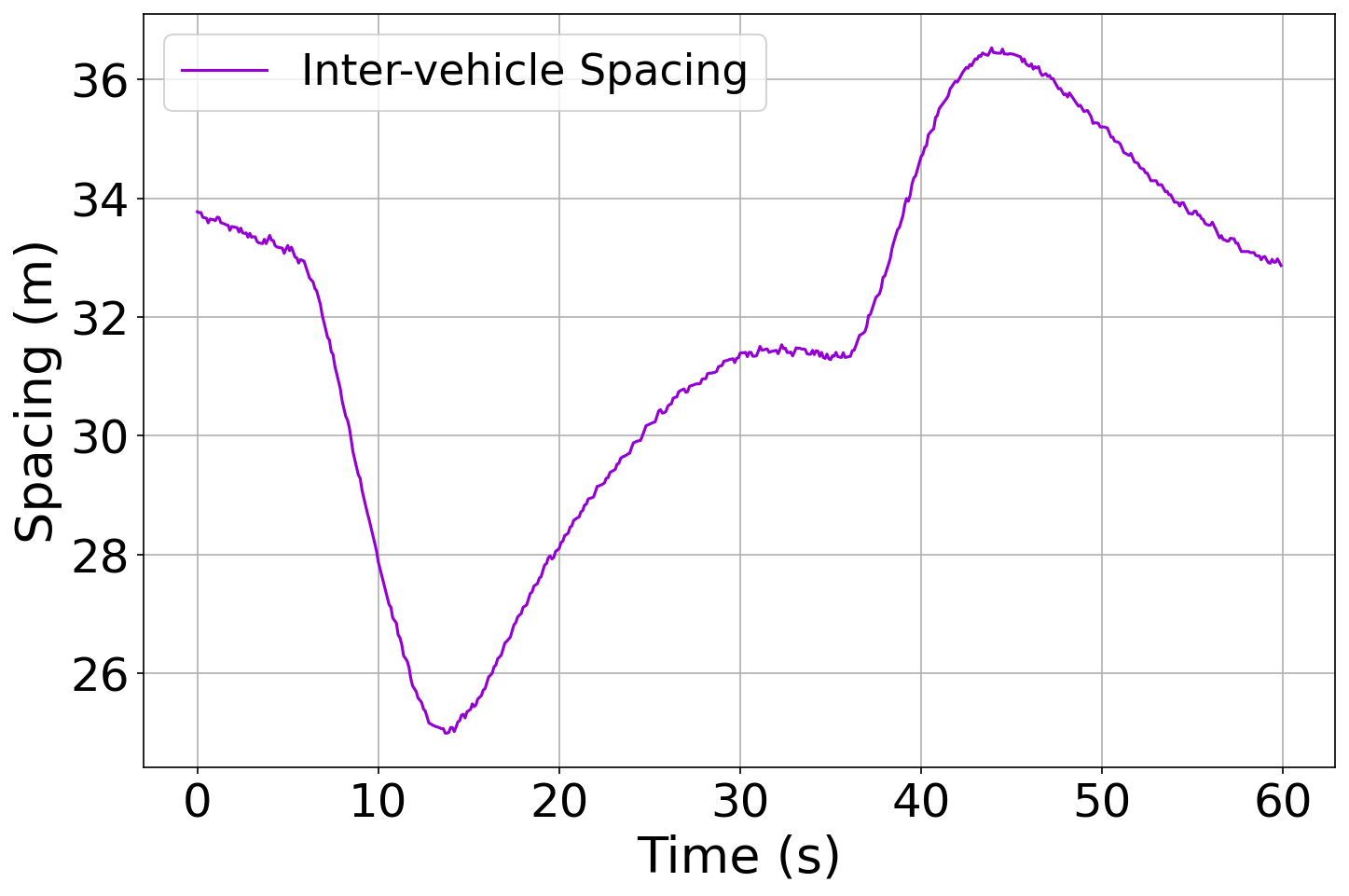}%
\label{fig_second_case}}
\newline 
\subfloat[Lead vehicle and follow vehicle speed (max setting).]{\includegraphics[scale = 0.14]{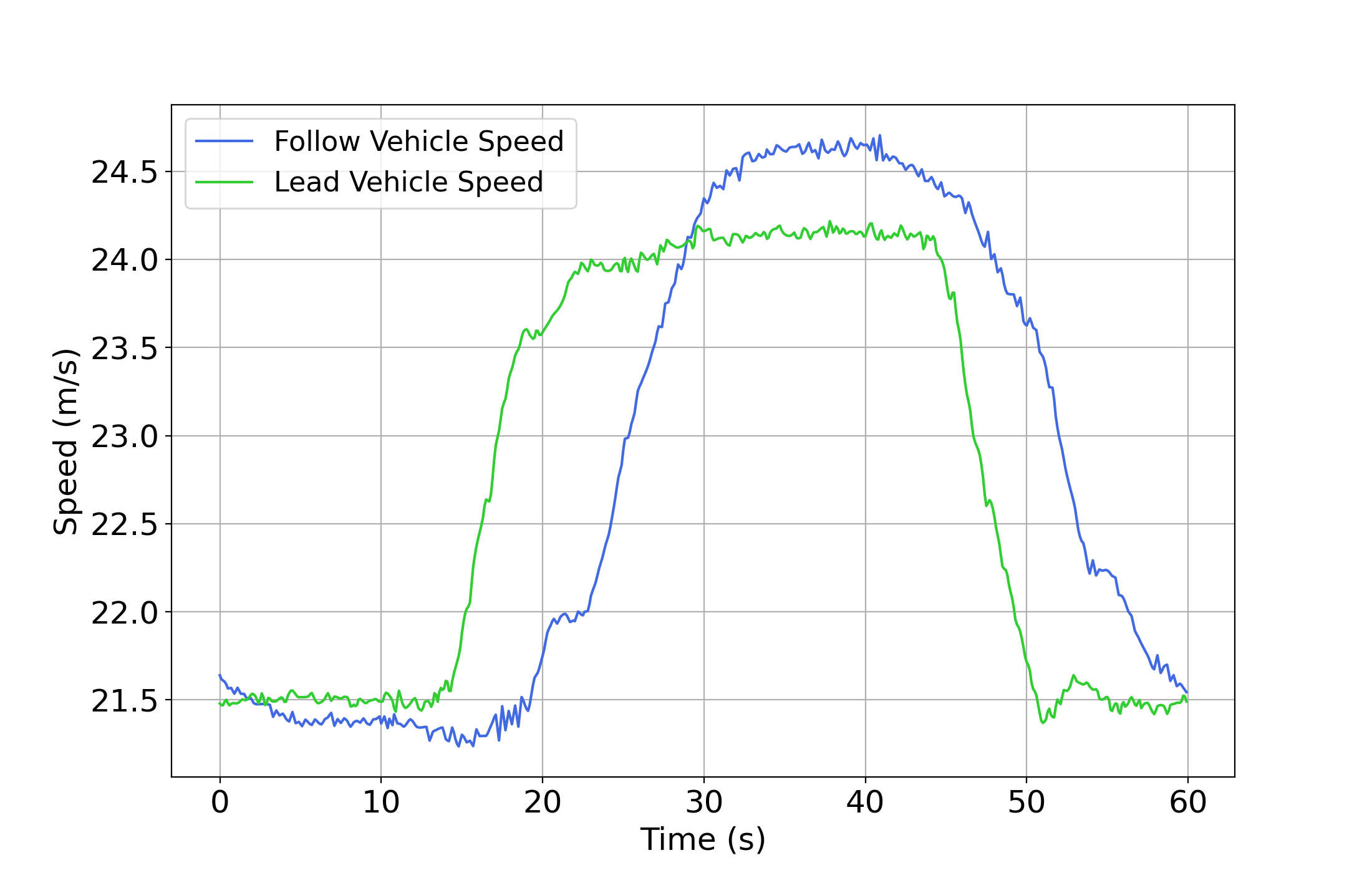}%
\label{fig_third_case}}
\hspace*{\fill} 
\subfloat[Inter-vehicle spacing (max setting).]{\includegraphics[scale = 0.14]{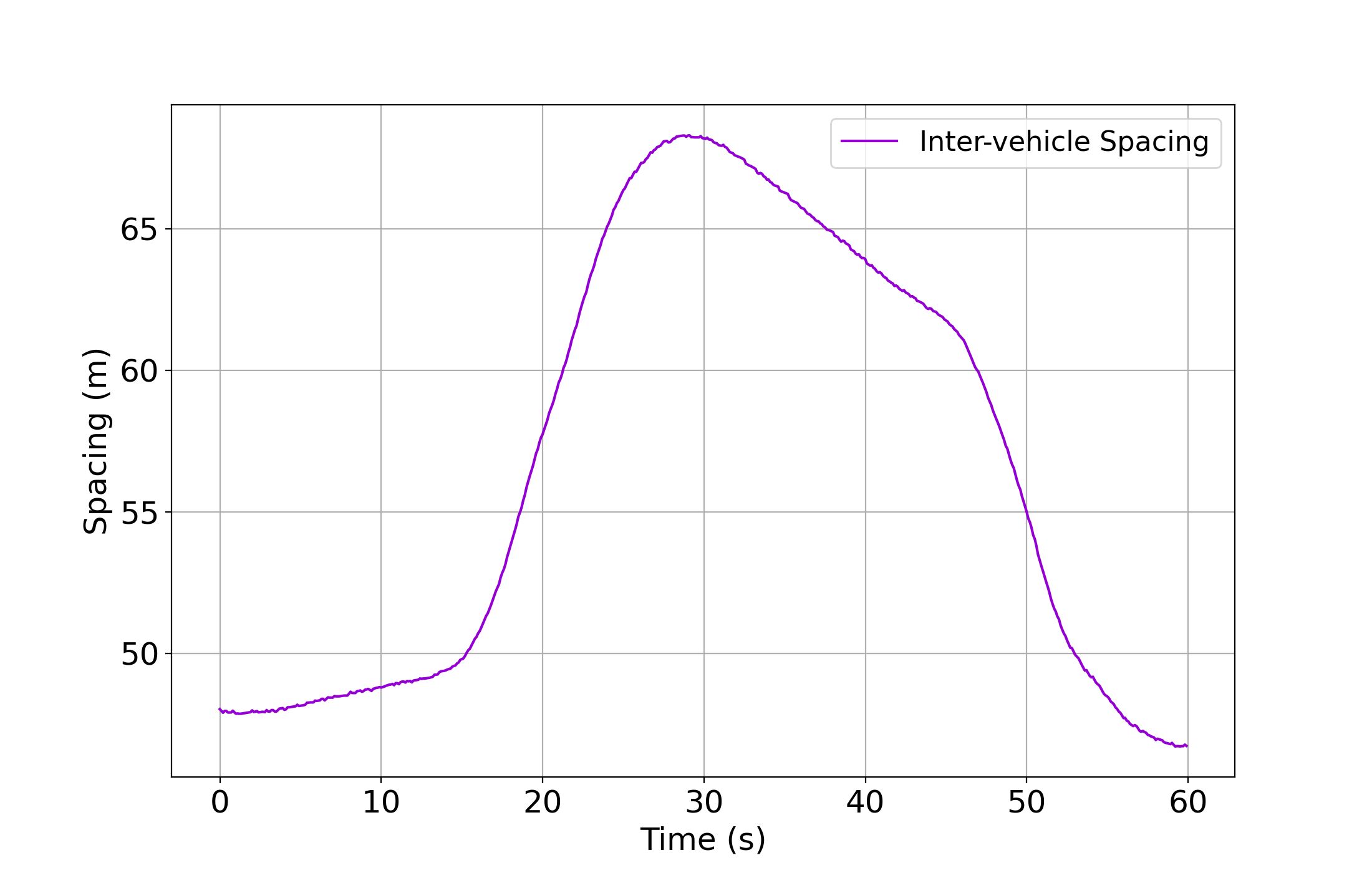}%
\label{fig_fourth_case}}
\caption{Comparison of vehicle trajectory data from the ACC car-following dataset under different ACC settings. (a) and (b) show the speed and spacing for the minimum gap ACC setting, while (c) and (d) show these for the maximum gap ACC setting.}
\label{fig:ACC_veh}
\end{figure}

To ensure accurate position and speed data, each vehicle is fitted with high-precision GPS receivers. This allows for the calculation of the inter-vehicle gap and relative speed. Fig.~\ref{fig:ACC_veh} provides a snapshot for one of the following vehicles under two conditions: minimum gap ACC setting, allowing the vehicle to maintain a close spacing to the one ahead, and maximum gap ACC setting, permitting greater spacing from the lead vehicle. This sample offers an illustrative example of leader-follower speed and spacing trajectories. The data collection operates at a 10 Hz sampling rate. Detailed information about the assorted set of vehicles utilized in Gunter et al.'s experiment~\cite{gunter2020are} is provided in Table~\ref{tab:vehicle_summary}. We used 80\% of the data for training, 10\% for validation, and the remaining 10\% for testing.

The ACC following vehicles are subjected to four distinct experimental conditions: oscillatory, low-speed steps, high-speed steps, and dips. The oscillatory trial is intended to assess the ACC system's behavior under fluctuating lead speed and headway conditions. The low-speed step trial aims to gather steady-state following behavior across an expansive range of low speeds. Similarly, the high-speed step trial collects steady-state driving trajectories at higher speeds. The dips trial evaluates the ACC vehicle's response to abrupt and substantial shifts in the lead vehicle's speed. 

\begin{table}
\renewcommand{\arraystretch}{1.3}
\caption{Summary of ACC vehicles tested by \cite{gunter2020are}.}
\label{tab:vehicle_summary} 
\centering
\begin{tabular}{ccccc}
\toprule
    Vehicle & Make & Style & Engine & \makecell{Min. Speed \\ (mph)}\\ 
    \midrule
    A & 1 & Full-size sedan & Combustion & 25\\
    B & 1 & Compact sedan & Combustion & 25\\
    C & 1 & Compact hatchback & Hybrid & 25\\
    D & 1 & Compact SUV & Combustion & 25\\
    E & 2 & Compact SUV & Combustion & 0\\
    F & 2 & Mid-size SUV & Combustion & 0 \\
    G & 2 & Full-size SUV & Combustion & 0\\
    \bottomrule
\end{tabular}
\end{table}

\section{Numerical Experiments}\label{sec:experiments}

This section presents the computational experiments undertaken to validate the efficacy of a rational neural network-based car-following model in car-following behavior modeling. We provide details of the experimental design, with a specific focus on the training, data assembly, and calibration of the physical parameters. The experiments are executed on a computing setup equipped with an Intel i7-12700K CPU @ 3.6 GHz processor complemented with 64 GB of memory. The deep learning-based models are trained and evaluated on a single NVIDIA GeForce RTX 3090 with 24 GB memory, utilizing the PyTorch 2.0 framework for these operations.

\subsection{Physical Model}

This study conducts experiments leveraging an experimental dataset collected by Gunter et al. during car-following trials with commercially available ACC vehicles, as discussed in Section~\ref{sec:data}. Specifically, our CF models are trained and validated using the oscillatory dataset. This dataset consists of car-following trajectories, where the lead vehicle executes pre-planned driving maneuvers while the trailing vehicle has its ACC activated. This precision of the GPS system used to collect the trajectory data allows for the measurement of the inter-vehicle space gap and relative speed with a fine granularity of 0.1 seconds.

To emulate car-following behavior precisely, the acceleration of the trailing vehicle is determined by evaluating its speed change over a time interval, corresponding to some time steps as the data is captured at a rate of 10 Hz (0.1 s intervals). The velocity of the trailing vehicle is represented as $V_f$. The equation for acceleration is:

\begin{equation}
\hat{a}(t) = \ddot{x}_f = \frac{V_f(t+0.1) - V_f(t)}{0.1}
\end{equation}

Here, $\ddot{x}_f$ represents the acceleration of the trailing vehicle. We use a time interval as an example, the velocities $V_f(t+0.1)$ and $V_f(t)$ represent the trailing vehicle's velocity at time $t+0.1s$ and $t$ respectively. The acceleration over the interval is the difference between these velocities divided by the time interval of 0.1 seconds.

In this study, we employ the OVRV, an extended model to the FVD Model~\cite{jiang2001full}. We select the OVRV model to depict the ACC vehicle driving behavior because it has demonstrated its ability to effectively encapsulate car-following dynamics of commercially-available ACC vehicles with a straightforward model~\cite{gunter2019modeling, shang2022novel, li2021classification}. Additionally, we have reviewed other relevant literature~\cite{punzo2021calibration, he2022physics} and implemented widely used physical models such as IDM and Gipps for comparison with the OVRV model, using the same data and calibration methods. For the Gipps model, which relies on speed control instead of acceleration control, we estimated acceleration using headway, as outlined in the literature~\cite{punzo2021calibration, he2022physics}, to ensure a fair comparison. The results show that the OVRV model performs best in both spacing and speed metrics, making it our chosen baseline physical model.

\begin{equation}
\ddot{x}_f = k_1(s-\eta-\tau v)+k_2\dot{s}
\end{equation}

The OVRV model is characterized by four model parameters that can be tuned to accurately represent car-following behavior. First, $k_1$ is the gain parameter on the constant effective time-gap term. Second, $k_2$ is a relative velocity parameter with respect to the lead vehicle. A higher $k_2$ value results in a quicker reaction to relative velocity changes. The $\tau$ parameter acts as a time constant, which is related to the inter-vehicle time gap. Finally, $\eta$ represents the jam distance (i.e., inter-vehicle spacing at zero speed). For this study, we use $k_1=0.052$, $k_2=0.236$, $\tau=0.796$, and $\eta=13.836$ for minimum gap setting, and $k_1=0.018$, $k_2=0.105$, $\tau=2.489$, and $\eta=0.0003$ for maximum gap setting through calibration.
The performance of the OVRV model, along with the simulation results, are presented in Section~\ref{sec:analysis}.

To calibrate the OVRV model, we minimize the MSE in inter-vehicle acceleration, using the equation:

\begin{equation}\label{eq:minCTH}
\begin{array}{rl}
\underset{\theta}{{\text{minimize}}}: & \sqrt{\frac{1}{N}\sum_0^N{({\ddot{x}}_\text{f}(t)-\hat{a}(t))^2}dt}\\
\text{subject to:}
& s(t) = \hat{s}(t)\\
& v_\text{l}(t) = \hat{v_\text{l}}(t)\\
& v_\text{f}(t) = \hat{v_\text{f}}(t)
\end{array}
\end{equation}

Here, the $\hat{s}(t)$, $\hat{v_\text{l}}$(t), and $\hat{v_\text{f}}$(t) represent real data, and $N$ denotes the number of data points in the training data. The optimization process in Equation~\ref{eq:minCTH} is solved via numerical optimization, ensuring that our model accurately predicts inter-vehicle acceleration based on the collected velocity data.

\subsection{Deep Learning CF Model}

In this follow-up experiment, we use the neural network modes in replicating the physical car-following behavior, specifically, we use the Long Short-Term Memory (LSTM) model, offering an alternative approach to modeling inter-vehicle dynamics. Incorporating the LSTM model into this approach is necessary because of its inherent capacity to retain long-term dependencies and accurately forecast future acceleration patterns of nonlinear data. LSTM has been proven to be capable of dealing with sequence data in several studies~\cite{graves2013hybrid,graves2013speech}, such as language process, and speech recognition, and shows that the LSTM model can model complex sequential interactions.
The LSTM has also been applied in the field of transportation in a variety of studies~\cite{cui2020stacked, li2022taxi, li2022detecting, mo2021physics}, including demand forecasting, car-following modeling, and trajectory reconstruction. Building on the success of our previous study that utilizes the OVRV model, we now delve into the realm of neural networks and deep learning techniques.

In our approach to crafting an accurate LSTM model, we consider acceleration calculations along with spacing, following velocity, and lead velocity as inputs. We establish an elaborate LSTM model architecture with five layers, each layer housing 64 LSTM units, complemented by an additional layer of dense connections. This meticulously designed architecture excels in capturing and learning the temporal dependencies that are intrinsic to the sequential car-following data, making it adept at modeling car-following dynamics. 

The functioning of the LSTM layer is based on the following equations:

\begin{equation}
f_{t} = \sigma(W_{f}\cdot [h_{t-1}, X_t] + b_{f})
\end{equation}

\begin{equation}
i_{t} = \sigma(W_{i}\cdot [h_{t-1}, X_t] + b_{i})
\end{equation}

\begin{equation}
\tilde{c}_{t} = \tanh(W{c}\cdot [h_{t-1}, X_t] + b_{c})
\end{equation}

\begin{equation}
c_{t} = f_{t} \odot c_{t-1} + i_{t} \odot \tilde{c}_{t}
\end{equation}

\begin{equation}
o_{t} = \sigma(W_{o}\cdot [h_{t-1}, X_t] + b_{o})
\end{equation}

\begin{equation}
h_{t} = o_{t} \odot \tanh(c_{t})
\end{equation}

In the above equations, $\sigma$ is the sigmoid activation function, $\odot$ denotes element-wise multiplication, $f_{t}$, $i_{t}$, $o_{t}$ are the forget, input, and output gates respectively, and $\tilde{c}_{t}$ is the candidate cell state. $W{f}$, $W_{i}$, $W_{o}$, and $W_{c}$ are the weight matrices and $b_{f}$, $b_{i}$, $b_{o}$, and $b_{c}$ are the bias terms. These equations illustrate the information flow through an LSTM unit at a given time step $t$, for a specific input $X_t$, and previous hidden state $h_{t-1}$ and cell state $c_{t-1}$. The output from the final LSTM layer and the last element of each input sequence are connected to a dense layer, leading to the formation of two parallel information streams. These streams are then merged and funneled through a final dense layer to generate the model output $y_{pred}$. To augment the model's robustness and accuracy, we deploy several essential strategies during the training phase. These include data shuffling to inhibit pattern memorization by the model and to bolster its generalization ability. We also adopt superior training practices such as early stopping and Z-score normalization to facilitate a more accurate and trustworthy representation of the original data. 

\subsection{PINN CF Model}
Combining deep learning and the OVRV model, we create the PINN model, inspired by~\cite{mo2021physics, naing2022dynamic}. Still, we use the OVRV as our physical model instead of the IDM model in their study since OVRV has been shown better for the ACC driving behavior modeling~\cite{shang2022novel, gunter2020are}. {Utilizing the same data as before, we edit the loss function of the deep learning model. 
From our previous experiment with the OVRV model, we use the most accurate calibration coefficients from the OVRV model from the previous model for $k_1,k_2,\tau,\eta$ respectively.} We then use the predictions from the OVRV model with these coefficients to create a predicted acceleration that could be compared to the prediction deep learning output. By using these two predictions, we can create a loss function for the neural network by using the following equation that takes in the root mean squared error of the true and predicted data of the neural $mse_{loss}$ and the physics error of the neural network prediction, which is the neural acceleration prediction minus the OVRV prediction $physics_{loss}$, with an alpha coefficient $\alpha$ weight that determines the balance between data discrepancy and physics discrepancy contributions:

\begin{equation}
Loss_{PINN} = \alpha \cdot MSE_{NN}+(1-\alpha) \cdot MSE_{Physics}
\end{equation}

During the model training process, we treat $\alpha$ as a hyper-parameter, enabling the neural network to actively modify its value and enhance the model's accuracy. Through iterative experimentation, we identify the $\alpha$ value that yields the optimal balance between minimizing data discrepancy and physics discrepancy. This approach leads to the development of a model with heightened accuracy.

\subsection{RACER Model}

The model proposed in this study leverages the same LSTM component as the deep learning and PINN models discussed earlier. However, as delineated in the methodology section, we utilize two distinct data inputs, $X_{\text{seq}}$ and $X_{\text{phy}}$. The employment of $X_{\text{phy}}$ data in conjunction with the modified model structure enables the computation of the predicted acceleration with respect to the input data comprising of velocity $v$, spacing $s$, and relative velocity $r$. Additionally, it allows us to enforce the RDCs during the model training process.

Our results demonstrate that the proposed model excels over the previously mentioned models in all performance metrics, including the predicted acceleration, and simulated spacing and speed, as gauged through simulation. This underscores the efficacy of our novel approach and paves the way for further improvements in modeling ACC driving behavior. The loss function deployed in this model is outlined in Equation~\eqref{exp_rdc_loss}, offering a balanced consideration of both the data and RDCs aspects of the model.

\begin{equation}
\begin{aligned}
Loss_{RDCs} &= MSE_{NN} + \lambda \cdot MSE_{RDCs}
\\
&= MSE(a_{pred}, a_{true}) + \lambda_1 \cdot RDC_{speed} 
\\
&+ \lambda_2 \cdot RDC_{spacing} + \lambda_3 \cdot RDC_{\text{relative speed}}
\label{exp_rdc_loss}
\end{aligned}
\end{equation}

\section{Analysis of the Numerical Experiments}~\label{sec:analysis}

This section presents an analysis of numerical experiments conducted on four distinct models: the OVRV model, the NN model, the PINN model, and the RACER model. These models are evaluated using test data, which includes inter-vehicle spacing, following velocity, and leading velocity variables. The performance of these models is scrutinized using data from ACC-equipped vehicles under both minimum and maximum gap settings, as previously described in Section~\ref{sec:data}.

\subsection{Model Evaluation: Use the trained model as CF controller}
As recommended by Punzo and Montanino~\cite{punzo2016speed}, we prefer to compare model performance using the cumulative inter-vehicle spacing rather than the instantaneous values or speed error. Consequently, we take into account the cumulative error for the temporal evolution of states.
Thus, to evaluate the model's performance, we use the trained or calibrated model as a car-following controller, and using our models' acceleration predictions, we reconstruct position and velocity trajectories. These trajectories (spacing and speed profiles) for the following vehicles are derived based on kinematic dynamics, using a time step of $\Delta t = 0.1$~s—aligned with the experimentally gathered data:

\begin{equation}\label{eq: simulation process}
\begin{bmatrix}
s \\
v
\end{bmatrix}_{t+\Delta t} =
\begin{bmatrix}
s \\
v
\end{bmatrix}_t 
+ 
\begin{bmatrix}
v_{l}-v \\
a_{pred}
\end{bmatrix}_t \Delta t
\end{equation}

In this study, we compare the generated trajectories with actual trajectories to assess discrepancies. To quantify errors in acceleration, inter-vehicle spacing, and velocity, we employ the root mean squared error (RMSE) metric. It's important to note that $a_{simulation}$, $s_{simulation}$, and $v_{simulation}$ represent the simulated acceleration, inter-vehicle spacing, and velocity, respectively. Conversely, $a_{actual}$, $s_{actual}$, and $v_{actual}$ denote the corresponding actual measurements obtained from real-world data. The discrepancies are quantified as follows:

\begin{equation}
RMSE_{acceleration} = \sqrt{\frac{1}{N}\sum_{i=1}^{N}(a_{simulation,i} - a_{actual,i})^2},
\end{equation}

\begin{equation}
RMSE_{spacing} = \sqrt{\frac{1}{N}\sum_{i=1}^{N}(s_{simulation,i} - s_{actual,i})^2},
\end{equation}

\begin{equation}
RMSE_{velocity} = \sqrt{\frac{1}{N}\sum_{i=1}^{N}(v_{simulation,i} - v_{actual,i})^2}.
\end{equation}

\begin{table}
\centering
\caption{Root mean squared errors for different models used as car-following controller with minimum gap setting.}
    \begin{tabular}{lcccc}
        \toprule
        & \textbf{RACER} & \textbf{OVRV} & \textbf{NN} & \textbf{PINN} \\
        \midrule
    \textbf{Acceleration (m/s$^2$)} & \bf 0.204 & 0.208 & 0.209 & 0.207 \\
    \textbf{Speed (m/s)} & \bf 0.09 & 0.17 & 0.125 & 0.114 \\
    \textbf{Spacing (m)} & \bf 0.261 & 1.47 & 0.305 & 0.272 \\
    \bottomrule
\end{tabular}
\label{tab:rmse_results}
\end{table}

\subsection{Model Evaluation: ACC Vehicle with Minimum gap Setting}

Table~\ref{tab:rmse_results} presents the RMSE for four evaluated models using data from the oscillatory experiment on Vehicle A under minimum and maximum gap settings. The RACER model demonstrates superior accuracy across all metrics, achieving the lowest RMSE in acceleration prediction, where all models perform comparably.

\begin{figure}
\centering
\includegraphics[scale=0.35]{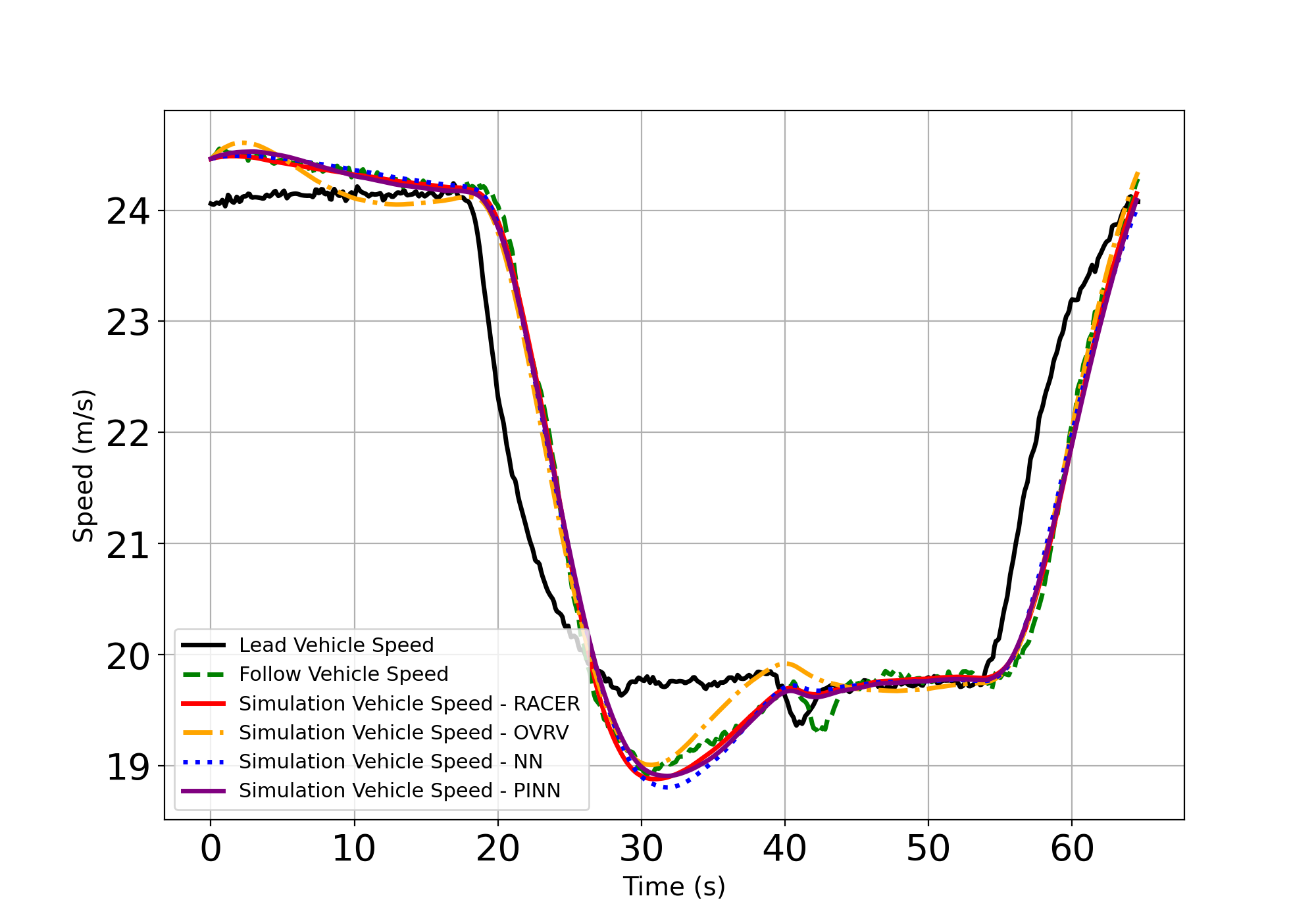}
\caption{Comparison of vehicle velocity from different simulation models over time.}
\label{fig:speed_compare}
\end{figure}

\begin{figure}
\centering
\includegraphics[scale=0.35]{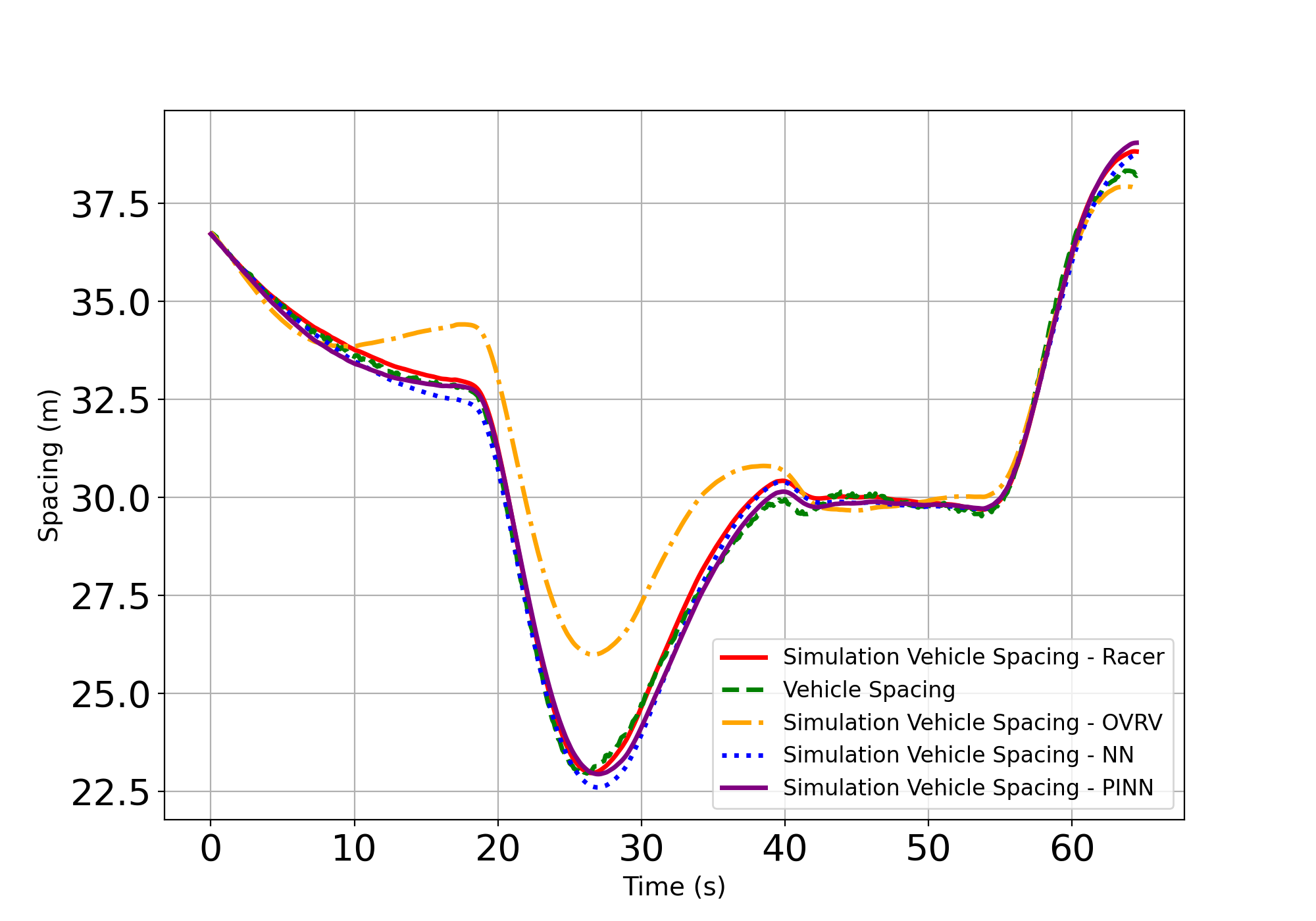}
\caption{Comparison of vehicle spacing from different simulation models over time.}
\label{fig:spacing_compare}
\end{figure}

The performance of these models extends to predicting speed and spacing, which are crucial parameters that depend on the leading vehicle's behavior. The OVRV model exhibits the weakest performance; its error margins for speed are at least twice as high as those of the RACER model and deteriorate further for spacing. Conversely, the PINN and NN models predict vehicle spacing with greater precision, as shown in Fig.~\ref{fig:spacing_compare}, and adapt more effectively to changes in the leading vehicle's speed, shown in Fig.~\ref{fig:speed_compare}. Nonetheless, the PINN and NN models' performance in speed and spacing prediction is not as consistent as the RACER model, with RMSE values exceeding those of the RACER model, indicating room for improvement.

\begin{figure}
\centering
\includegraphics[scale=0.35]{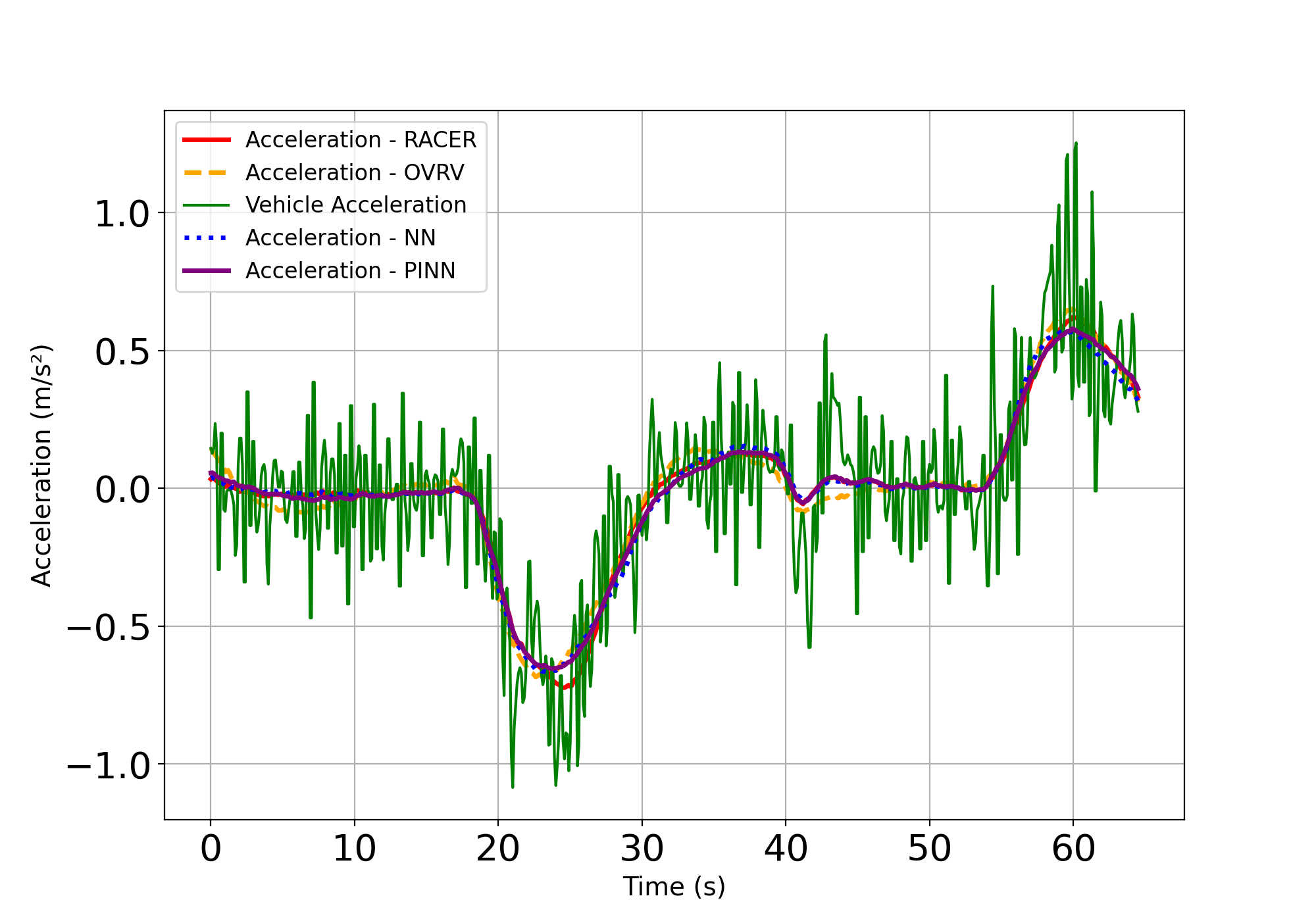}
\caption{Comparison of vehicle acceleration predictions from different simulation models over time.}
\label{fig:accel_compare}
\end{figure}

\begin{figure*}
\centering
\subfloat[][RDC Violations - Speed]{\includegraphics[scale = 0.23]{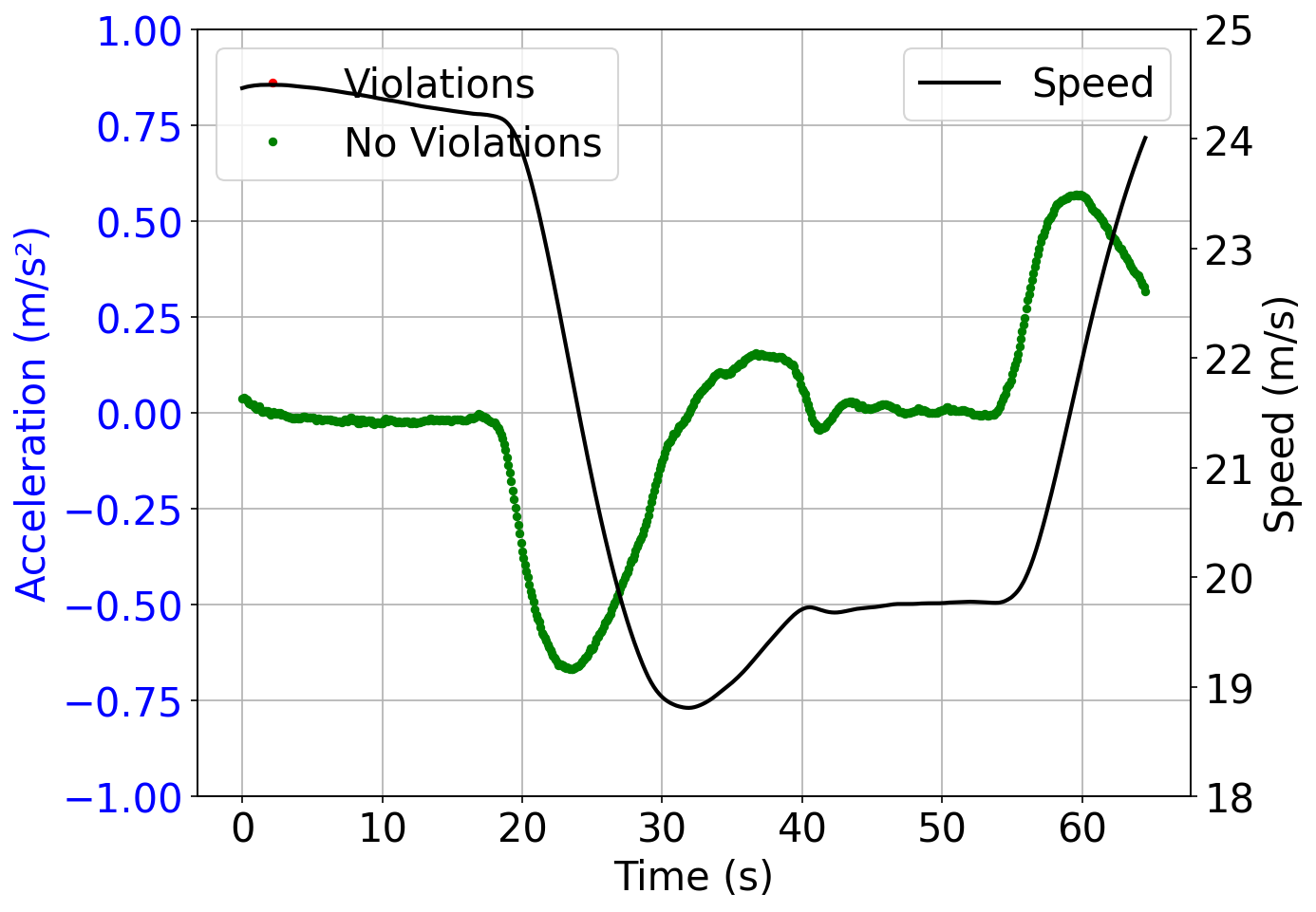}\label{rdc_1_1}}
\subfloat[][RDC Violations - Spacing]{\includegraphics[scale = 0.23]{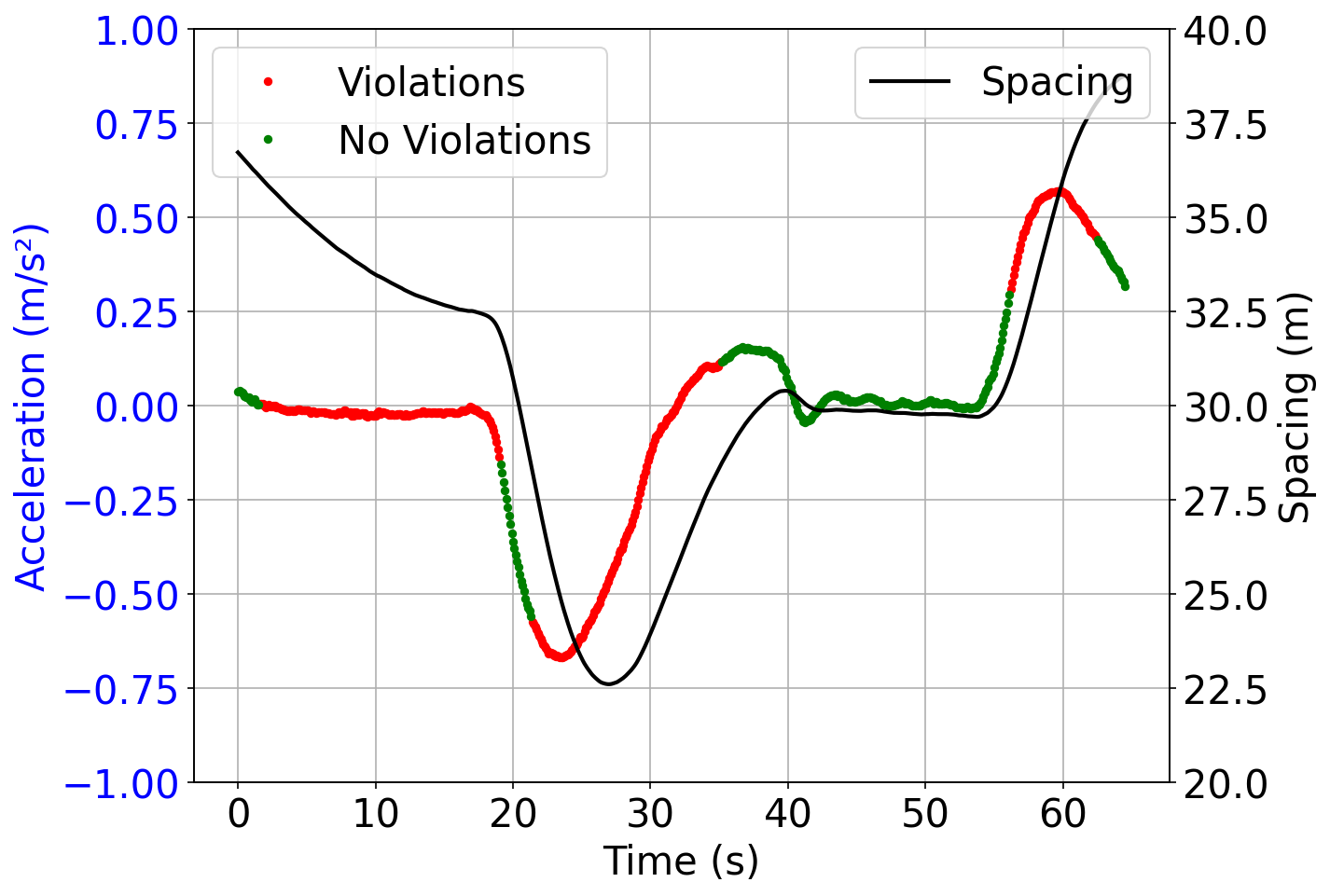}\label{rdc_1_2}}
\subfloat[][RDC Violations - Relative Speed]{\includegraphics[scale = 0.23]{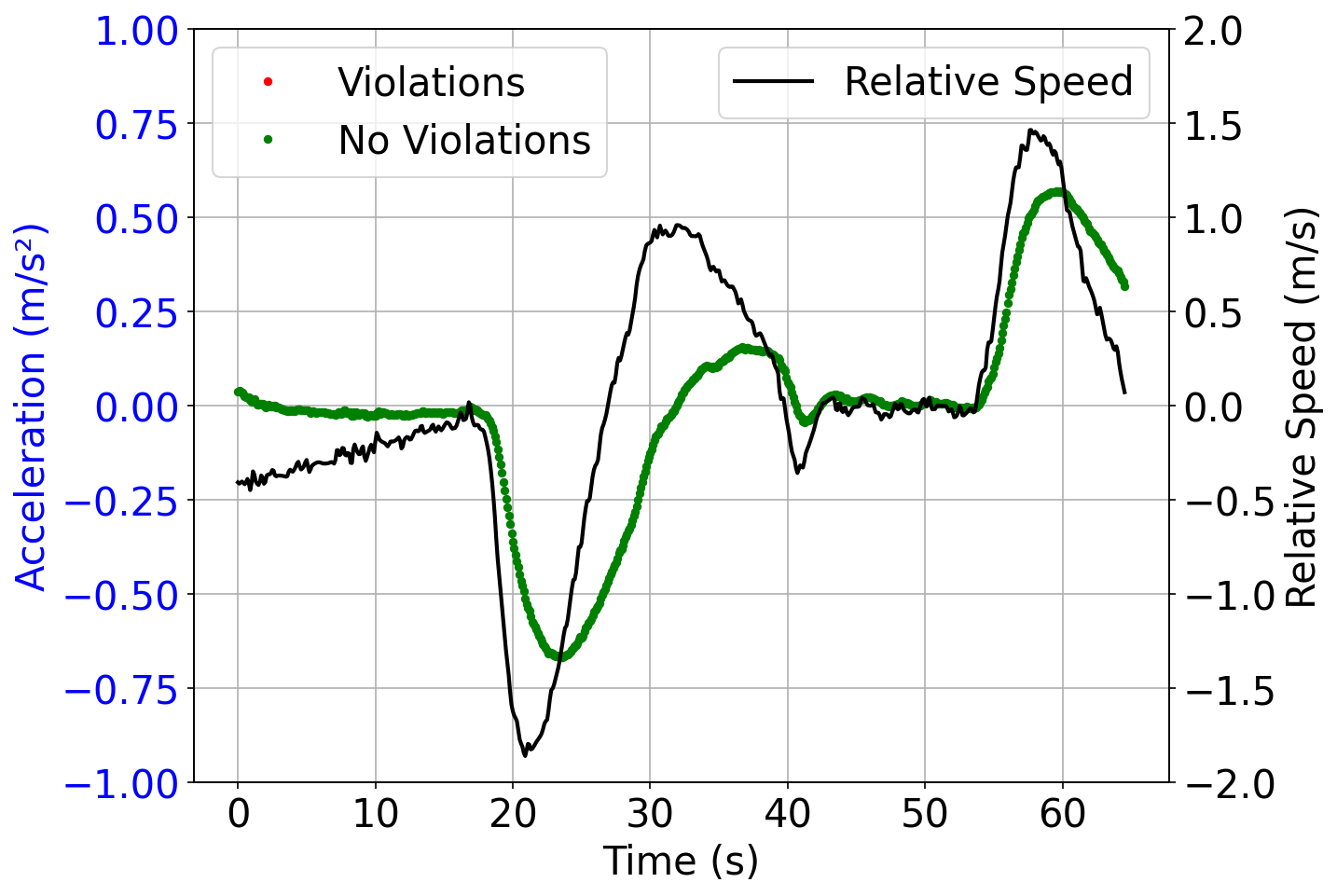}\label{rdc_1_3}}
\caption{Illustrations of predictions and violations by the LSTM Neural Network model in terms of speed, spacing, and relative speed. The green and red dots denote predictions conforming to and violating the established rules, respectively.}
\label{fig:RDC_NN}
\end{figure*}

\begin{figure*}
\centering
\subfloat[][RDCs Violations - Speed]{\includegraphics[scale = 0.23]{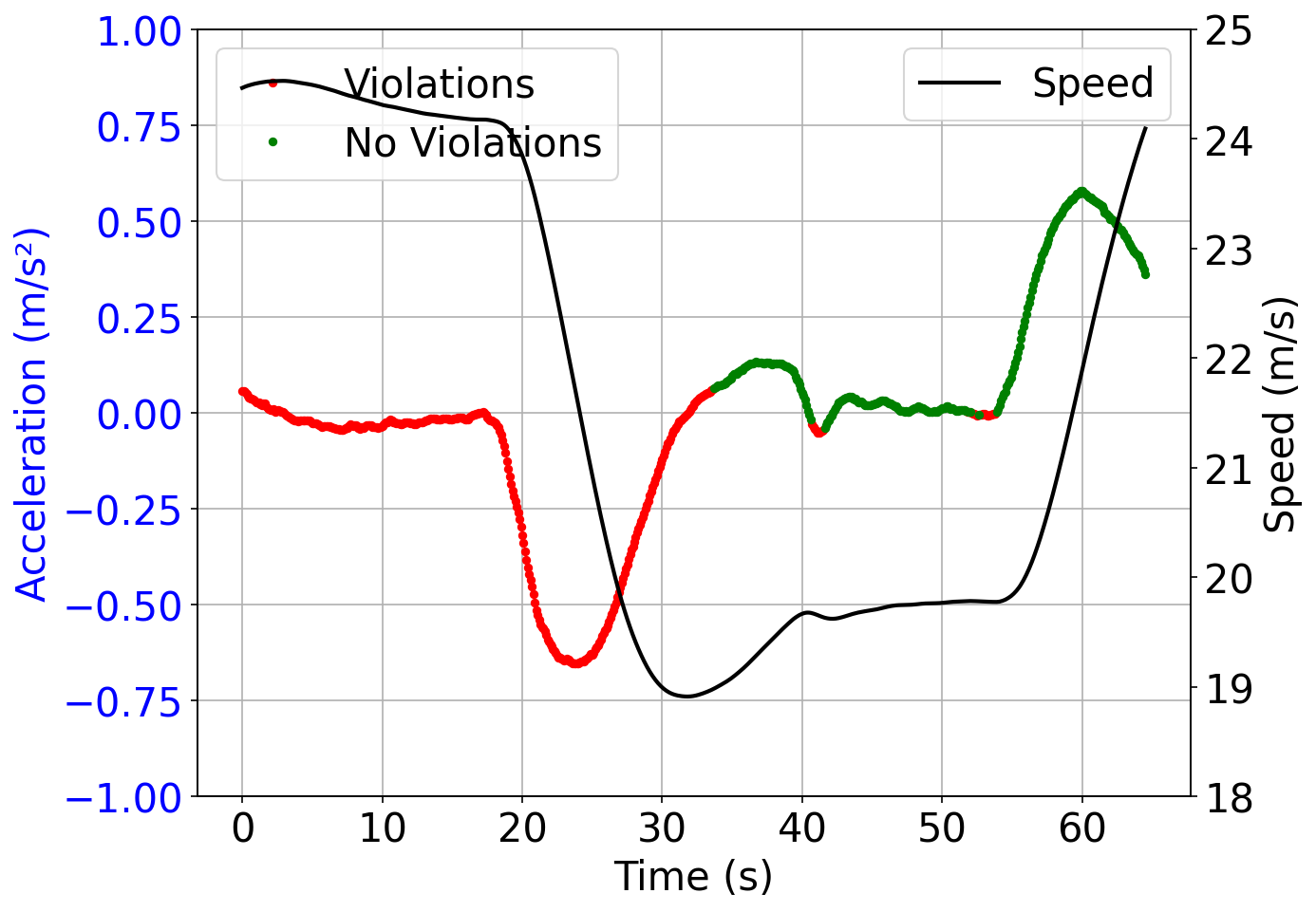}\label{rdc_2_1}}
\subfloat[][RDCs Violations - Spacing]{\includegraphics[scale = 0.23]{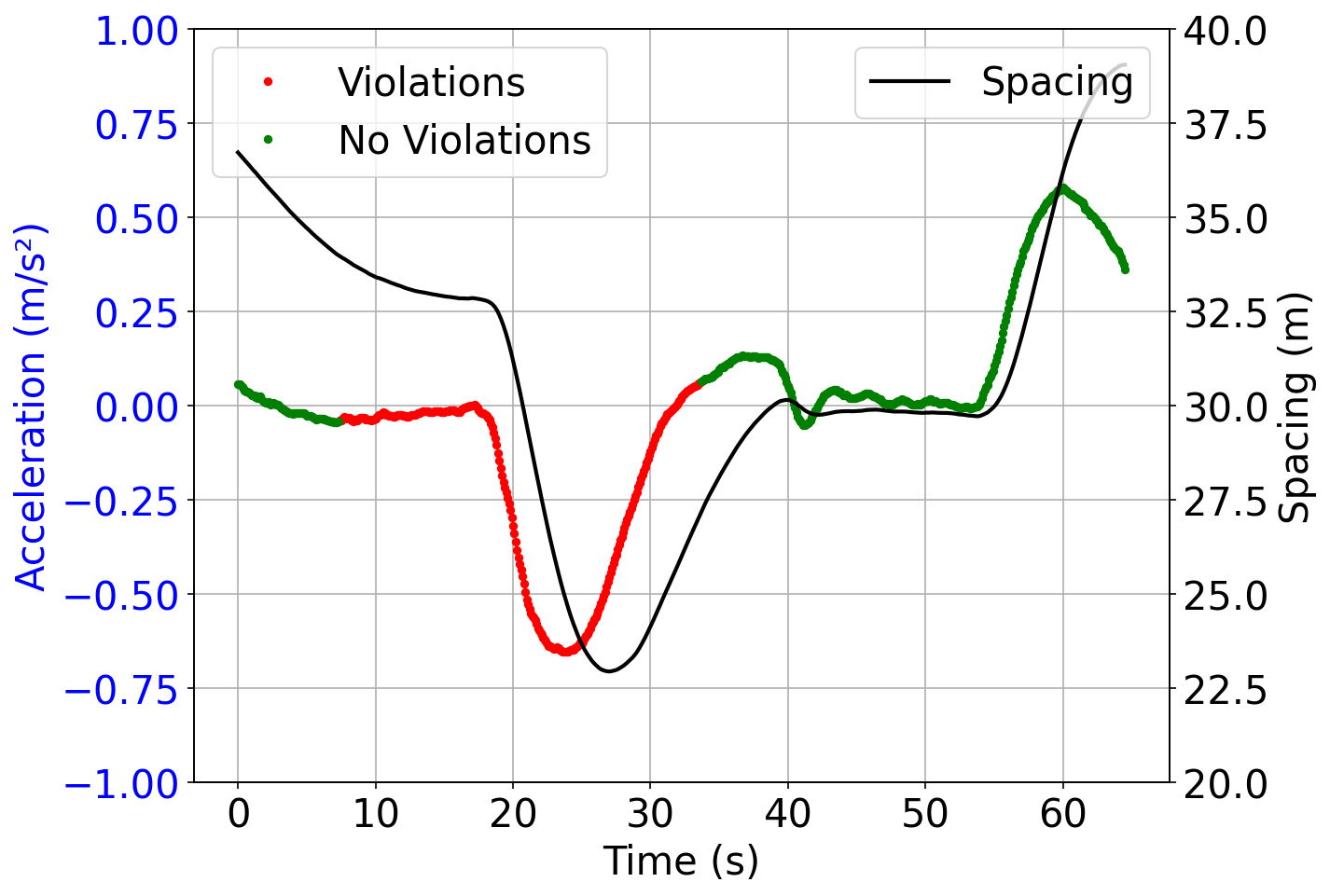}\label{rdc_2_2}}
\subfloat[][RDCs Violations - Relative Speed]{\includegraphics[scale = 0.23]{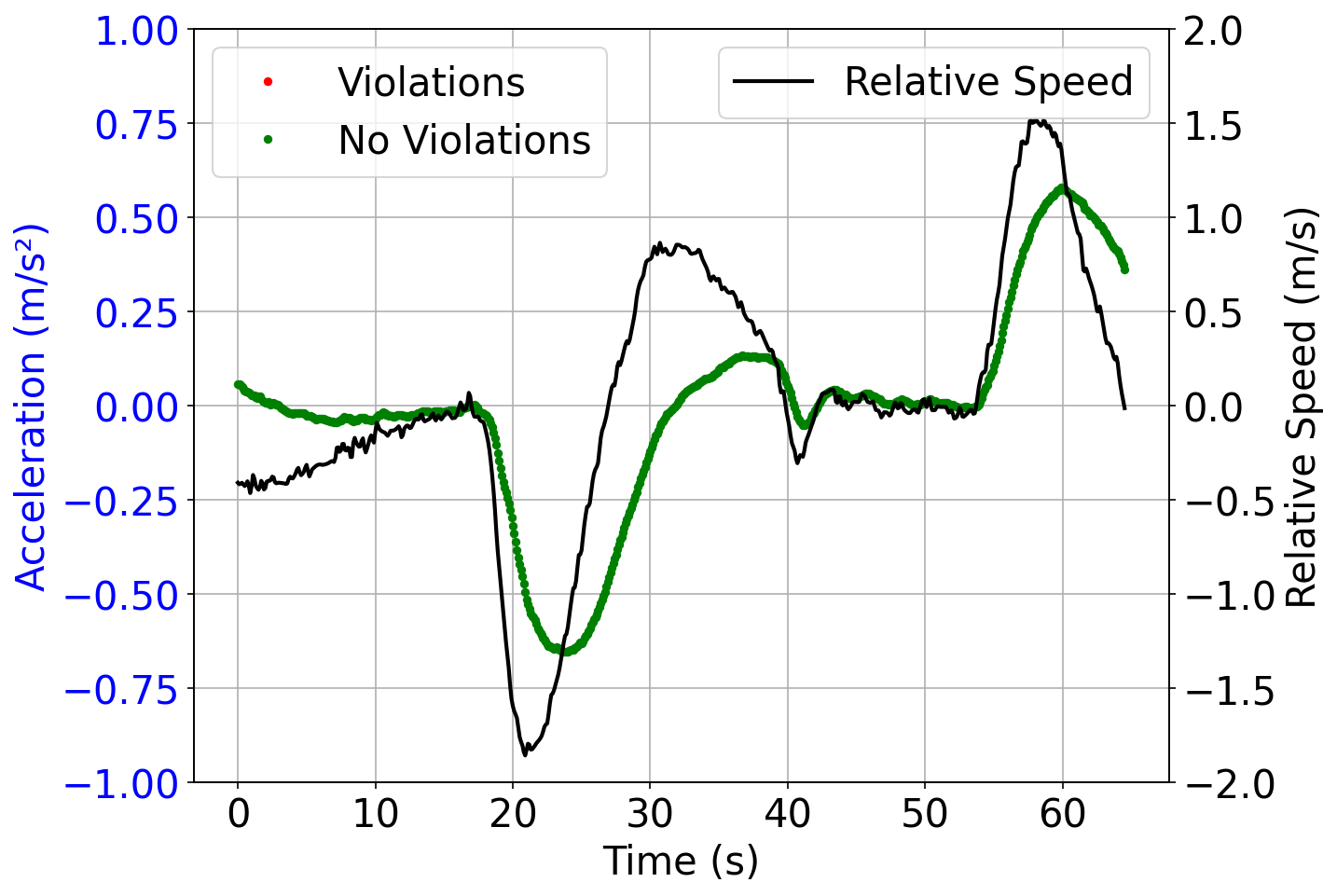}\label{rdc_2_3}}
\caption{Illustrations of predictions and violations by the physics-informed Neural Network model in terms of speed, spacing, and relative speed. The green and red dots denote predictions conforming to and violating the established rules, respectively.}
\label{fig:RDC_PINN}
\end{figure*}

\begin{figure*}
\centering
\subfloat[][RDC Violations - Speed]{\includegraphics[scale = 0.23]{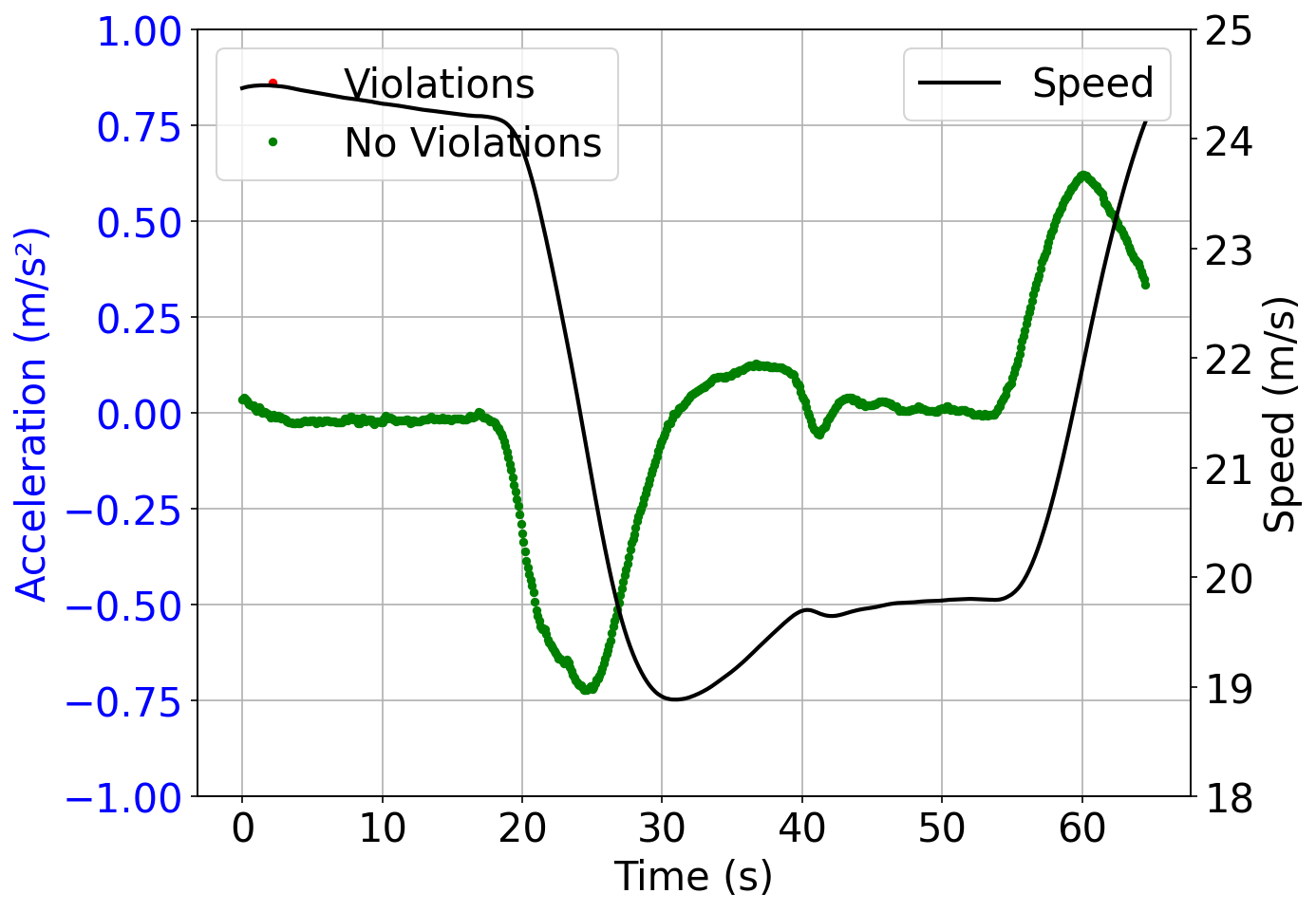}\label{rdc_3_1}}
\subfloat[][RDC Violations - Spacing]{\includegraphics[scale = 0.23]{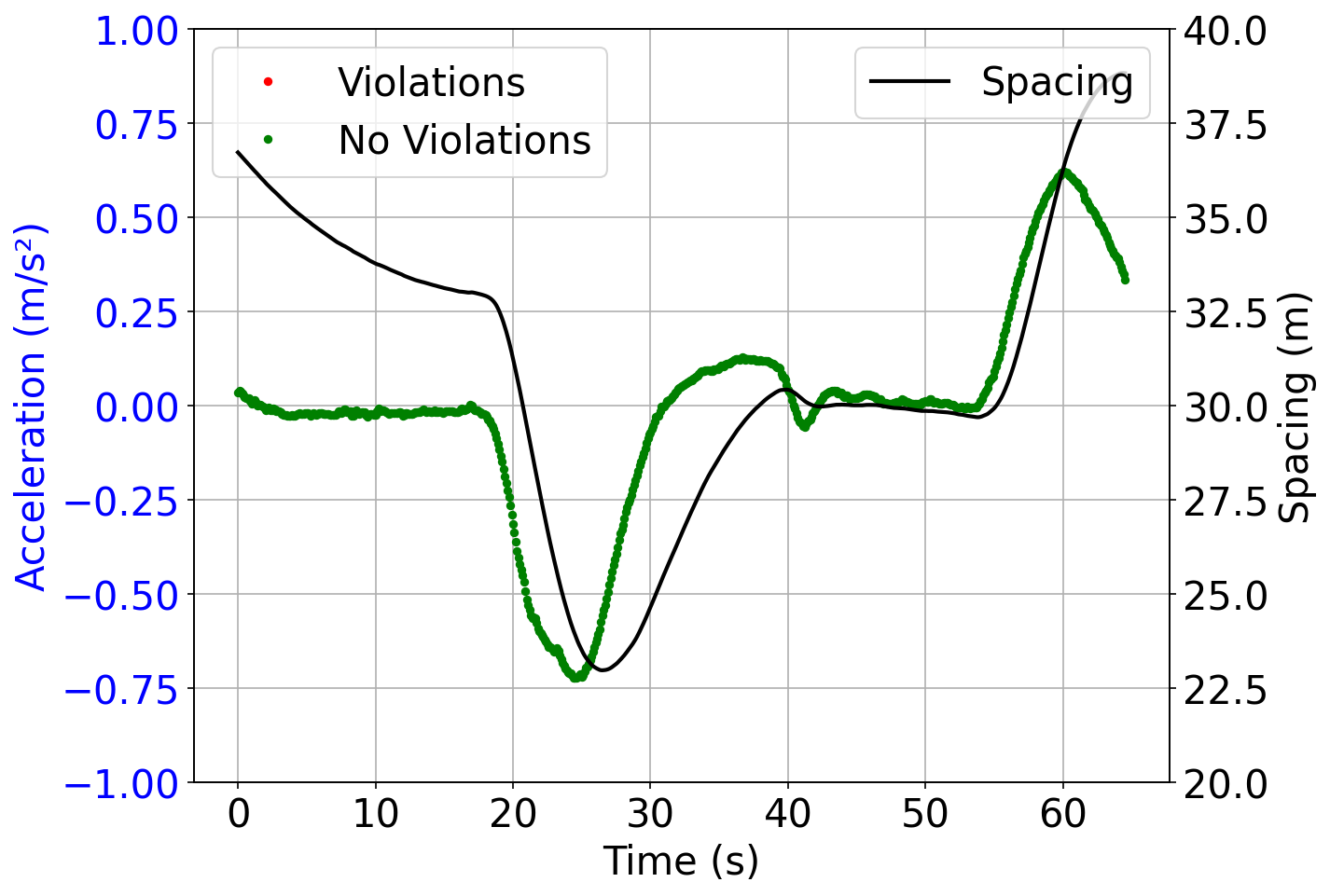}\label{rdc_3_2}}
\subfloat[][RDC Violations - Relative Speed]{\includegraphics[scale = 0.23]{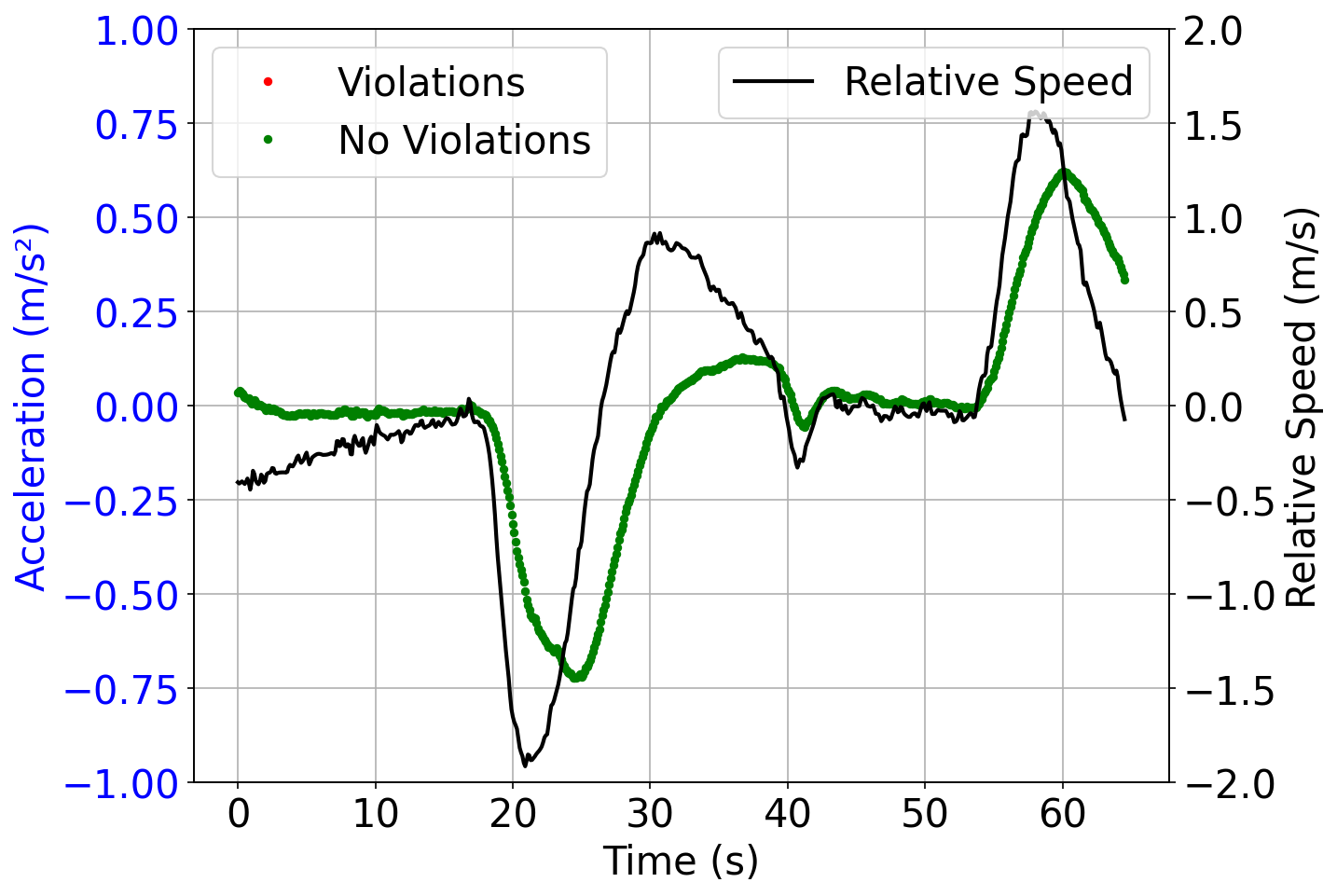}\label{rdc_3_3}}
\caption{Illustrations of predictions and violations by the RACER model in terms of speed, spacing, and relative speed. The green and red dots denote predictions conforming to and violating the established rules, respectively.}
\label{fig:RDC_NN_rational}
\end{figure*}

We then scrutinize the predictions made by various deep learning models against testing data, specifically focusing on compliance with the RDCs. These constraints evaluate the model-predicted accelerations to three driving parameters: velocity ($v$), spacing ($s$), and relative speed ($r$). To accurately compute these constraints, we utilize PyTorch's automatic differentiation function, which allows us to compute the precise gradients of acceleration with respect to each variable. This approach ensures that the derivatives are computed exactly at each time step, rather than approximating them numerically. As such, any violations in the RDCs are detected based on the multivariate derivatives across velocity, spacing, and relative speed.

Fig.~\ref{fig:RDC_NN} presents the results for a baseline NN model. Intriguingly, the NN managed to fulfill the RDCs relative to speed and relative speed constraints but fell short in the context of spacing, exhibiting violations approximately 60\% of the time, respectively. These violations are highlighted in the figure using red lines to indicate where the partial derivative $\frac{da}{dv} \leq 0$, signaling a violation of the RDC. Additionally, the model's performance with respect to velocity did not improve significantly, despite some minor non-violation segments.

A different perspective is provided in Fig.~\ref{fig:RDC_PINN}, which shows the performance of a PINN. Although the PINN model also complied fully with the RDCs of relative speed constraint, it exhibited a stark contrast when it came to velocity and spacing constraints, with an approximately 50\% violation rate for both these parameters.

Finally, we turn our attention to our proposed model, the Rational Neural Network, the results of which are delineated in Fig.~\ref{fig:RDC_NN_rational}. Using the same automatic differentiation approach for detecting RDC violations, this network demonstrates flawless performance, with zero RDC violations for relative speed, velocity, or spacing. Unlike the other models, this network successfully adheres to the constraints across all variables. This outcome highlights the efficacy of our model in adhering to real-world driving principles, further reinforcing the benefits of integrating domain-specific knowledge into deep learning models for more realistic and reliable predictions. The superior performance of RACER can be attributed to three key mechanisms: (i) the explicit enforcement of RDCs prevents the model from learning physically implausible behaviors that lead to prediction errors, (ii) the dual-input architecture ($X_{\text{seq}}$ and $X_{\text{phy}}$) allows the model to leverage both temporal patterns and instantaneous physical constraints simultaneously, and (iii) the gradient-based RDC enforcement through automatic differentiation ensures that the learned mapping respects fundamental driving principles at every prediction step, leading to more stable and realistic long-term trajectory simulation.

The OVRV model's poor performance in spacing prediction (RMSE of 1.47 m vs. RACER's 0.261 m) stems from its simplified linear relationships that cannot capture the complex, nonlinear dynamics present in real ACC behavior. The standard Neural Network's high violation rates (~60\% for spacing constraints) occur because it optimizes purely for prediction accuracy without physical constraints, leading to mathematically optimal but physically implausible solutions that violate basic driving principles. The PINN model's moderate violations (~50\%) result from the soft enforcement of physics through the loss function—while it encourages physics compliance, it cannot guarantee it, allowing violations when data fitting objectives conflict with physical principles.

\subsection{Model Evaluation: ACC Vehicle with Maximum Setting}

\begin{table}
\centering
\caption{Root mean squared errors for different models used as car-following controllers with maximum gap setting.}
    \begin{tabular}{lcccc}
        \toprule
        & \textbf{RACER} & \textbf{OVRV} & \textbf{NN} & \textbf{PINN} \\
        \midrule
    \textbf{Acceleration (m/s$^2$)} & \bf 0.221 & 0.614 & N/A (crash) & 0.227 \\
    \textbf{Speed (m/s)} & \bf 0.081 & 1.839 & N/A (crash) & 0.268 \\
    \textbf{Spacing (m)} & \bf 0.394 & 22.761 & N/A (crash) & 2.573 \\
    \bottomrule
\end{tabular}
\label{tab:rmse_results_max}
\end{table}

\begin{figure*}
\centering
\subfloat[Vehicle acceleration from prediction.]{\includegraphics[scale = 0.3]{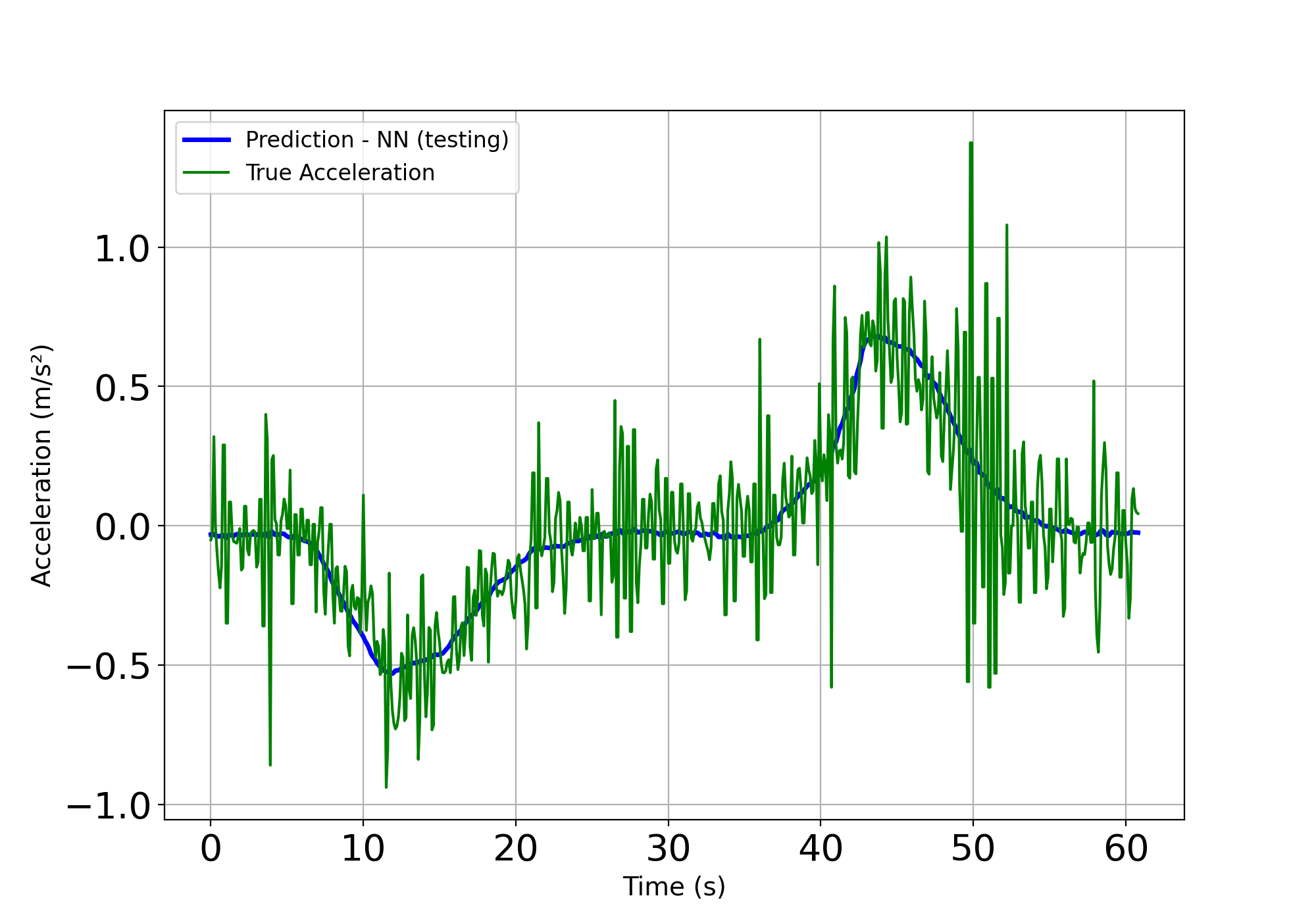}%
\label{fig:nn_pred}}
\hspace*{\fill}
\subfloat[Vehicle acceleration from simulation.]{\includegraphics[scale = 0.3]{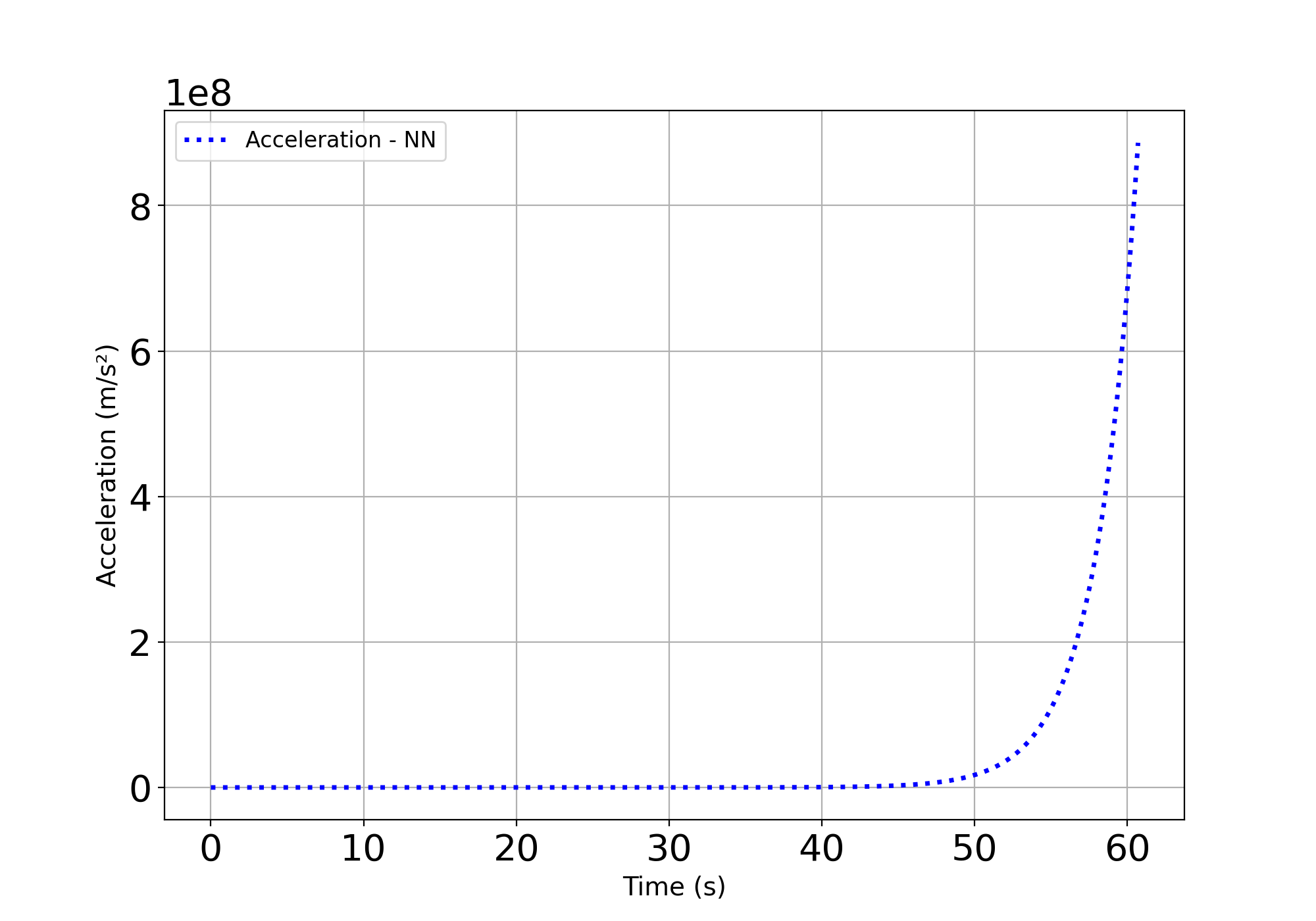}%
\label{fig:nn_sim}}
\caption{Vehicle acceleration of, a) predicted using testing data, and b) simulated using the trained model as a controller.}
\label{fig:nn_com}
\end{figure*}

The performance of the tested models under the maximum gap ACC setting diverges notably from that observed under the minimum gap setting. The RACER model maintains consistent efficacy, replicating its minimum gap setting proficiency as indicated in Table~\ref{tab:rmse_results_max}. The RACER model surpasses other models, especially in spacing and speed metrics. In particular, the performance of the NN model drastically declines in the maximum gap setting scenario, resulting in collisions during the simulation. In contrast, the NN model experiences a significant decline in performance under the maximum gap scenario, resulting in collisions during the simulation. While the NN model's predictions align well with the training data, as shown in Fig.~\ref{fig:nn_pred}, its application as a car-following controller leads to erratic and unstable behavior, as depicted in Fig.~\ref{fig:nn_sim}. This highlights a key insight: a model’s strong performance in standard testing does not necessarily translate to success in real-world scenarios. Specifically, in the context of car-following modeling, it is critical to validate a model through driving behavior simulations rather than relying solely on acceleration predictions based on test data. Given the NN model's collision incidents and subsequent simulation failure, it will be excluded from the following discussion. The RMSE for the NN model diverges significantly from the other models, indicating a disparity in predictive fidelity.

\begin{figure}
\centering
\includegraphics[scale=0.33]{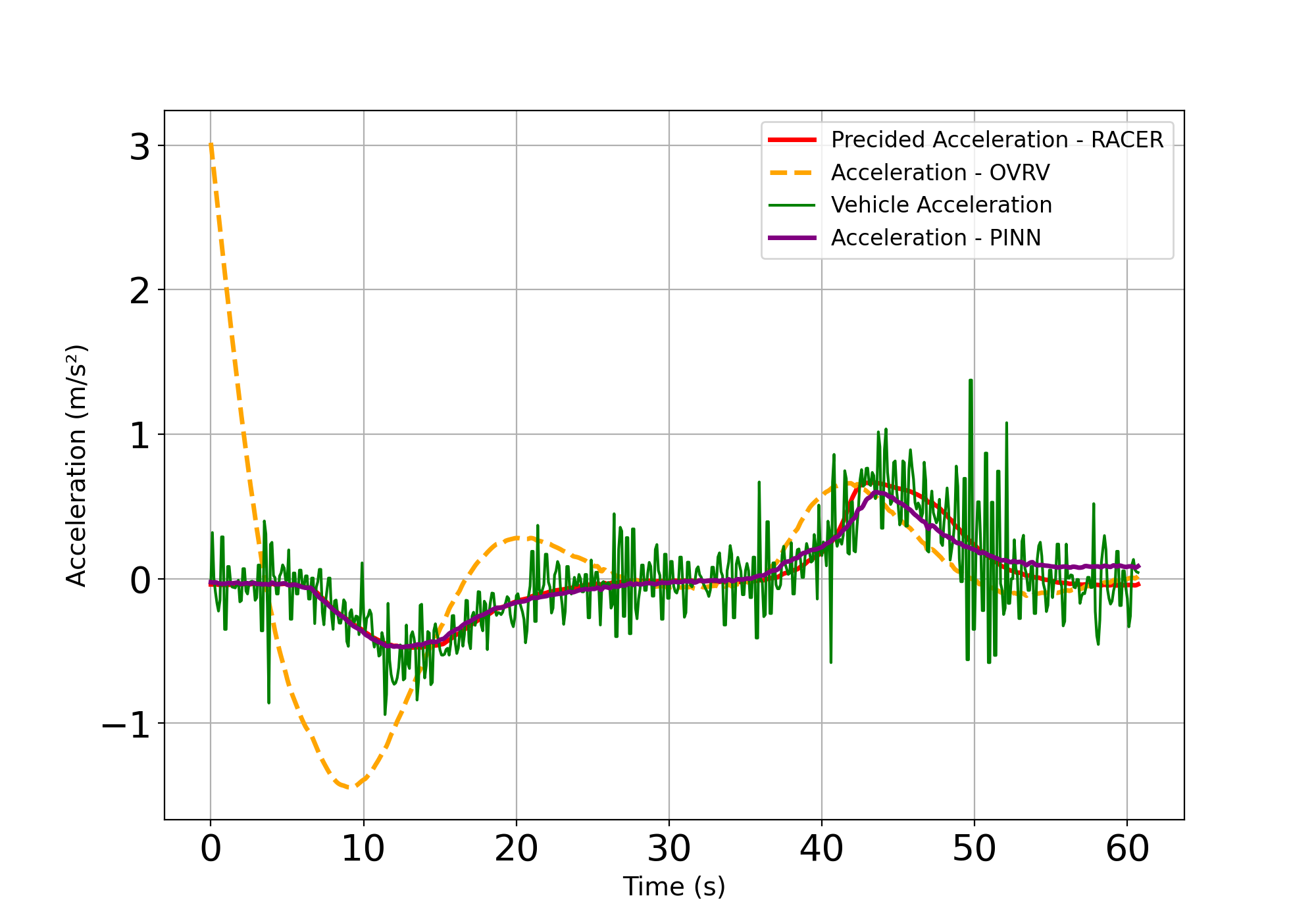}
\caption{Comparison of vehicle acceleration from different simulation models over time.}
\label{fig:accel_compare_max}
\end{figure}

For the maximum gap setting, the OVRV model's acceleration predictions have the highest RMSE (0.614 $m/s^2$), significantly exceeding the lowest recorded RMSE of 0.221 $m/s^2$ by the RACER model. The RMSE of the PINN model stands at a competitive 0.227 $m/s^2$, closely trailing RACER and demonstrating consistent performance across various driving conditions, including acceleration and deceleration phases, as shown in Fig.~\ref{fig:accel_compare_max}. Nonetheless, the speed and spacing predictions by both the PINN and OVRV models do not match the accuracy of the RACER model, with the spacing RMSE being approximately an order of magnitude greater and the speed RMSE at least twice as large. This discrepancy is a testament to the challenges posed by maximum gap setting conditions and suggests that model efficacy during training does not always translate to dependable control capabilities.

\begin{figure}
\centering
\includegraphics[scale=0.33]{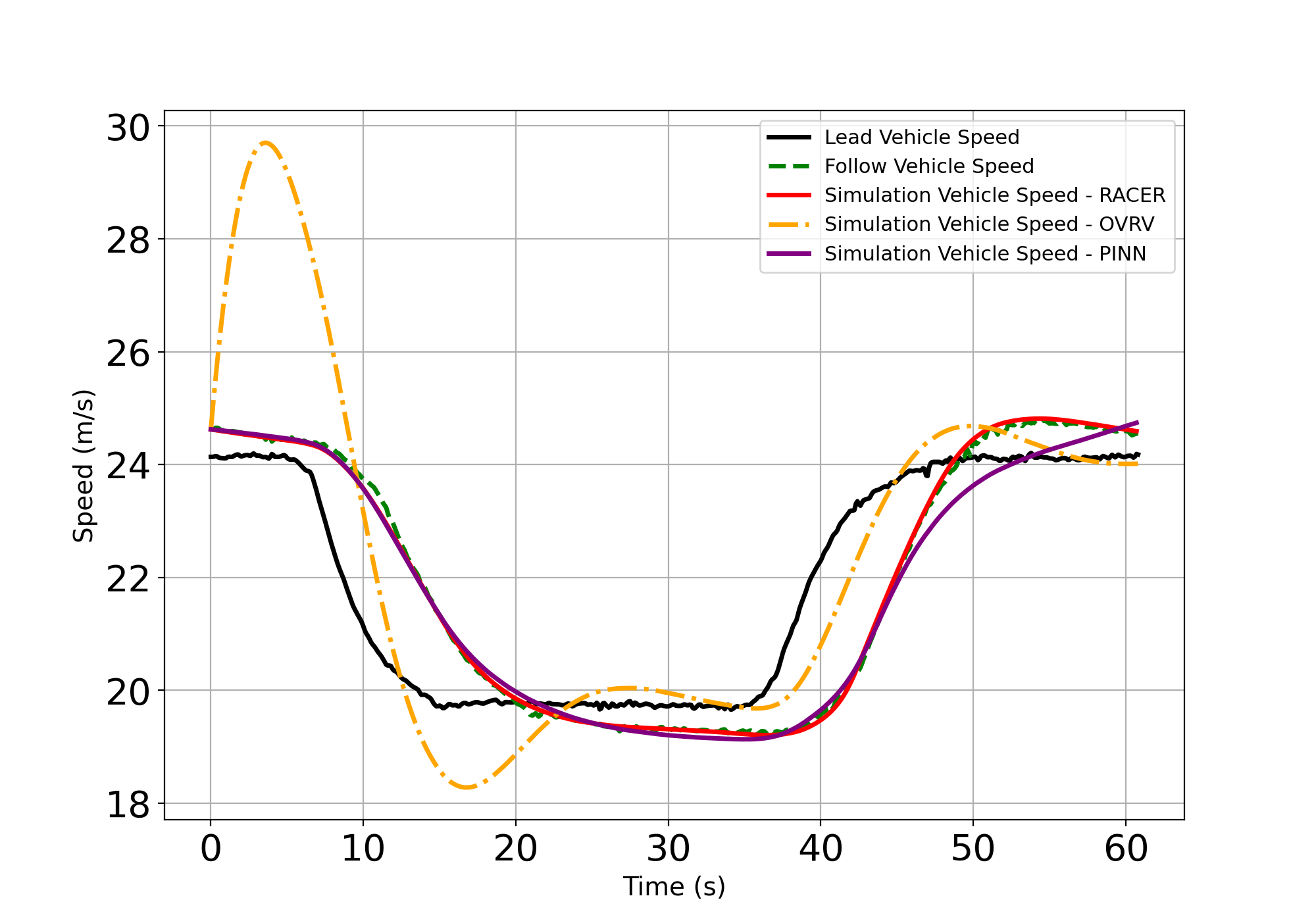}
\caption{Comparison of vehicle velocity from different simulation models over time.}
\label{fig:speed_compare_max}
\end{figure}

\begin{figure}
\centering
\includegraphics[scale=0.35]{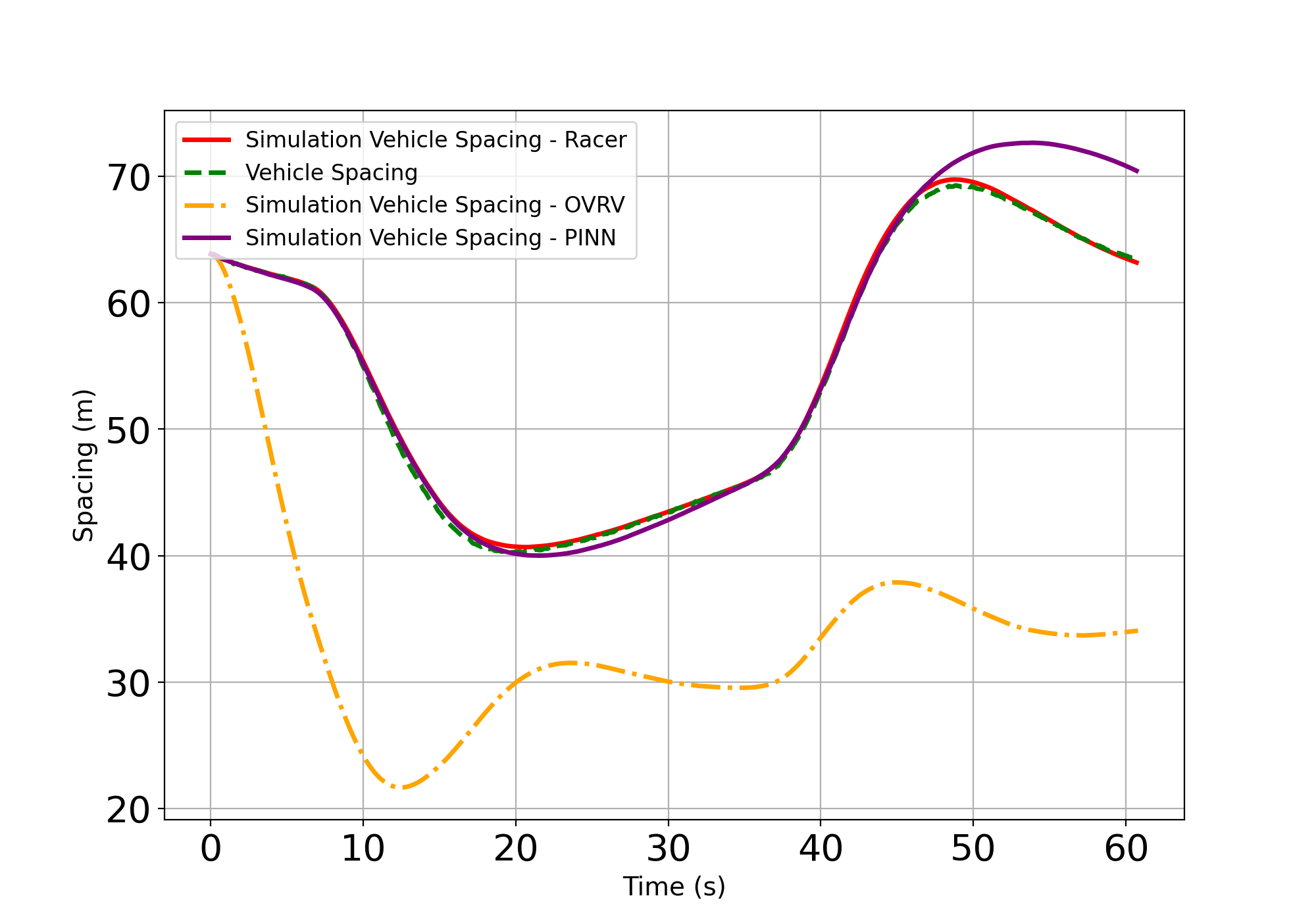}
\caption{Comparison of vehicle spacing from different simulation models over time.}
\label{fig:spacing_compare_max}
\end{figure}

As presented from Fig.~\ref{fig:accel_compare_max} through~\ref{fig:spacing_compare_max}, the OVRV model exhibits comparatively poor performance, particularly at the onset of the simulation. Notably, the spacing maintained by the OVRV model is less than that of both the RACER model and the originally observed spacing, raising potential safety concerns if used as a vehicle controller. The PINN model performs better than the OVRV model but still shows a divergence in the later stages of the simulation. This divergence contributes to its elevated RMSE in both speed and spacing metrics.

\begin{figure*}
\centering
\subfloat[][RDCs Violations - Speed]{\includegraphics[scale = 0.2]{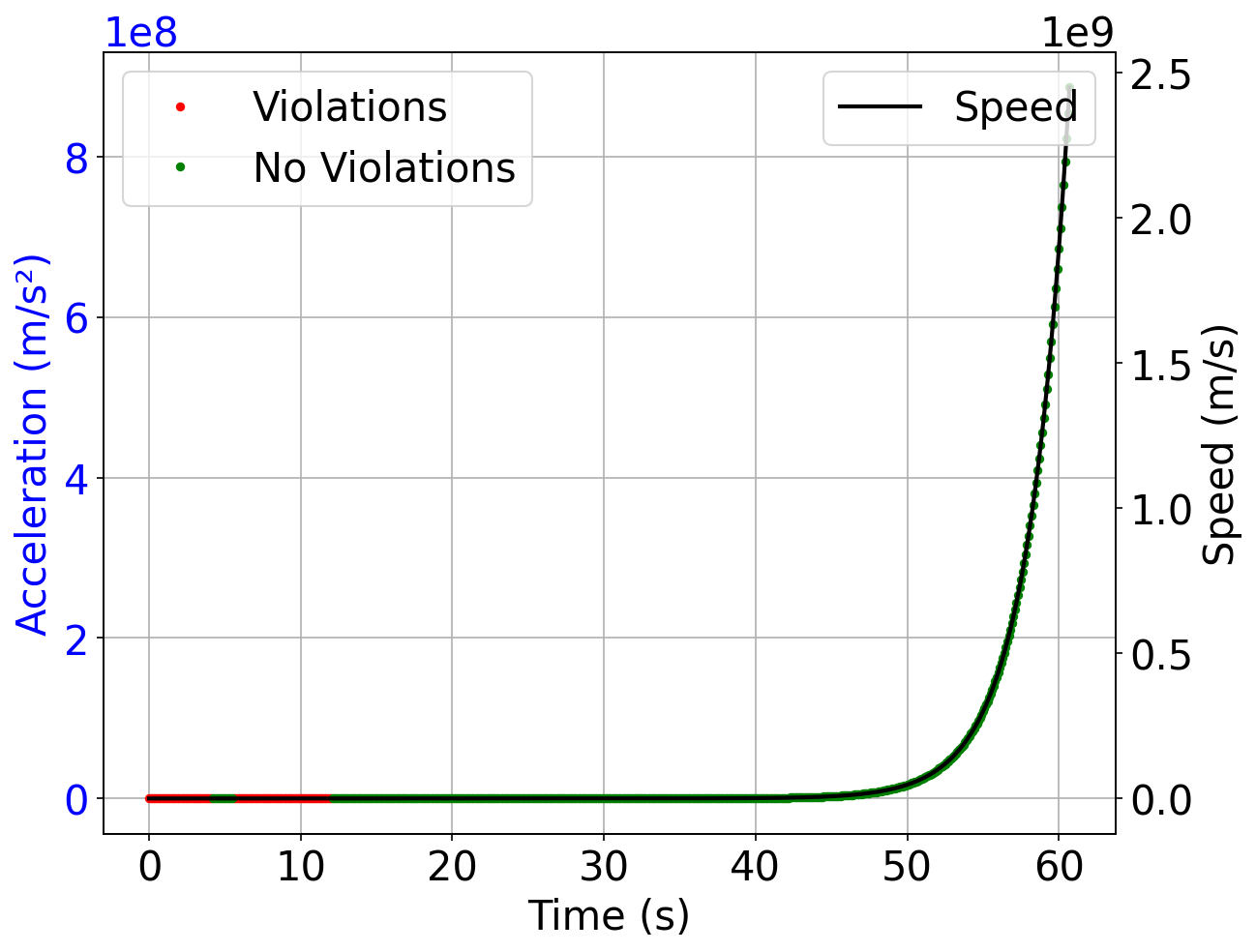}\label{rdc_1_1_max}}
\subfloat[][RDC Violations - Spacing]{\includegraphics[scale = 0.2]{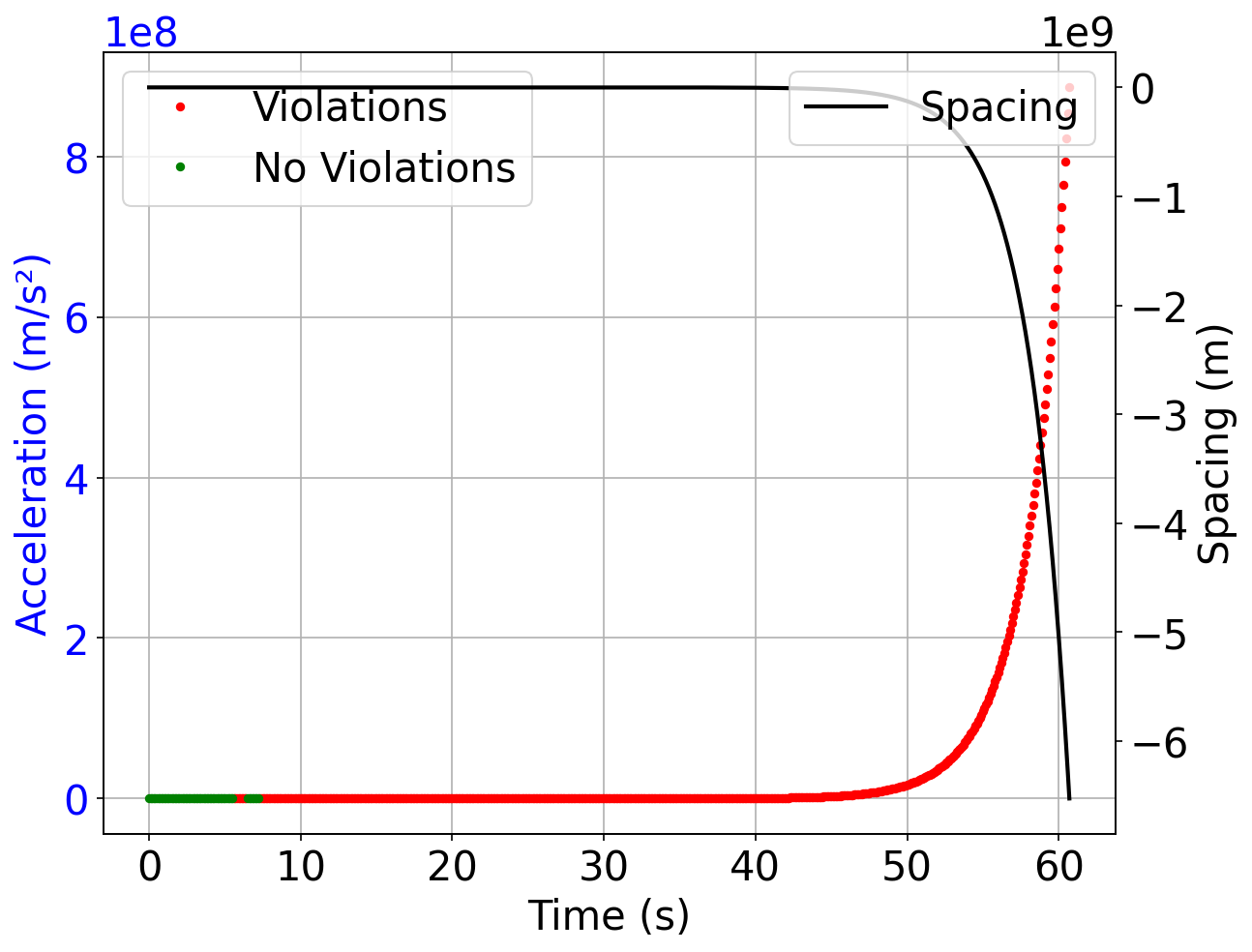}\label{rdc_1_2_max}}
\subfloat[][RDC Violations - Relative Speed]{\includegraphics[scale = 0.2]{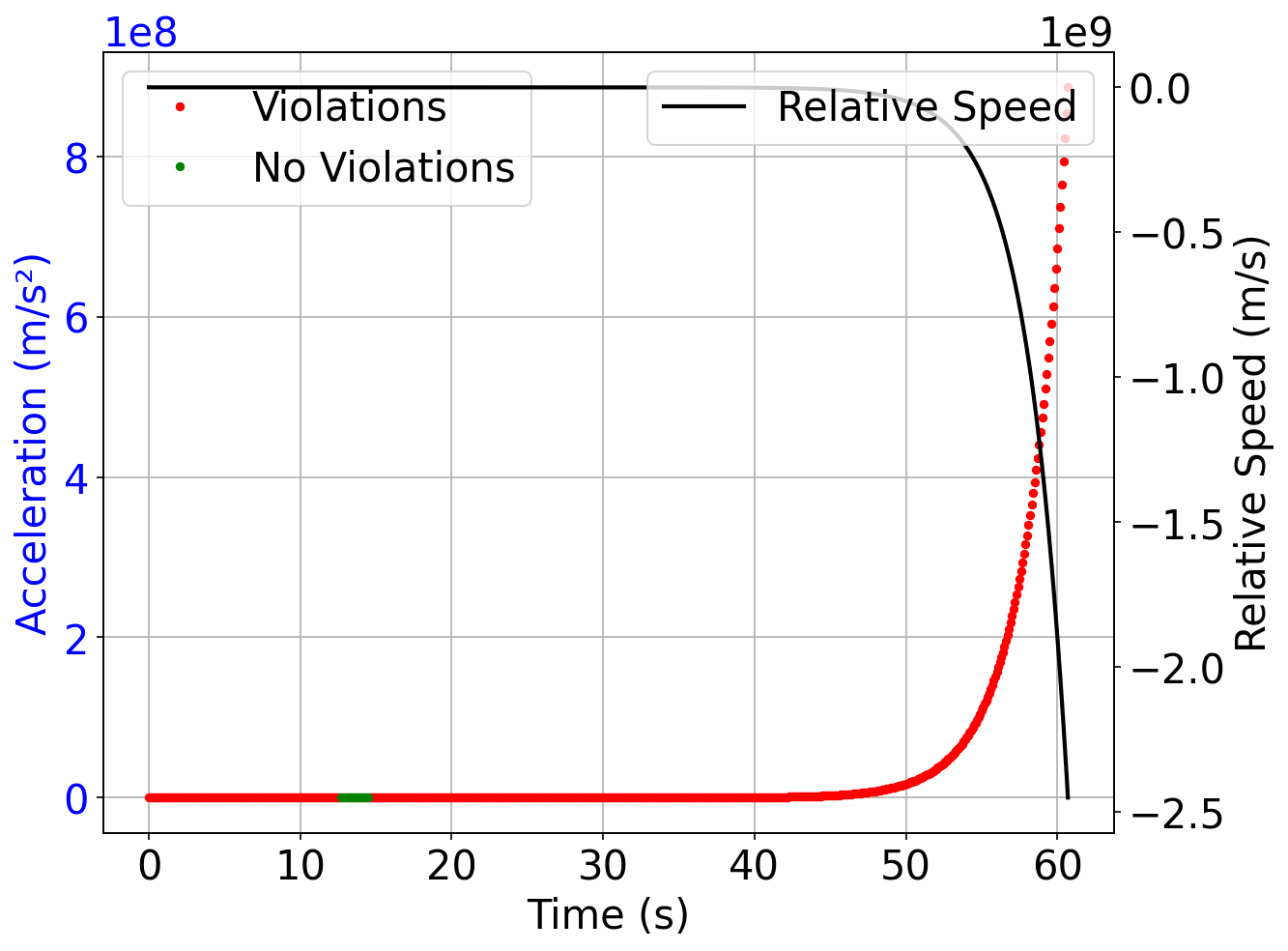}\label{rdc_1_3_max}}
\caption{Illustrations of predictions and violations by the LSTM Neural Network model in terms of speed, spacing, and relative speed. The green and red dots denote predictions conforming to and violating the established rules, respectively.}
\label{fig:RDC_NN_max}
\end{figure*}

\begin{figure*}
\centering
\subfloat[][RDCs Violations - Speed]{\includegraphics[scale = 0.2]{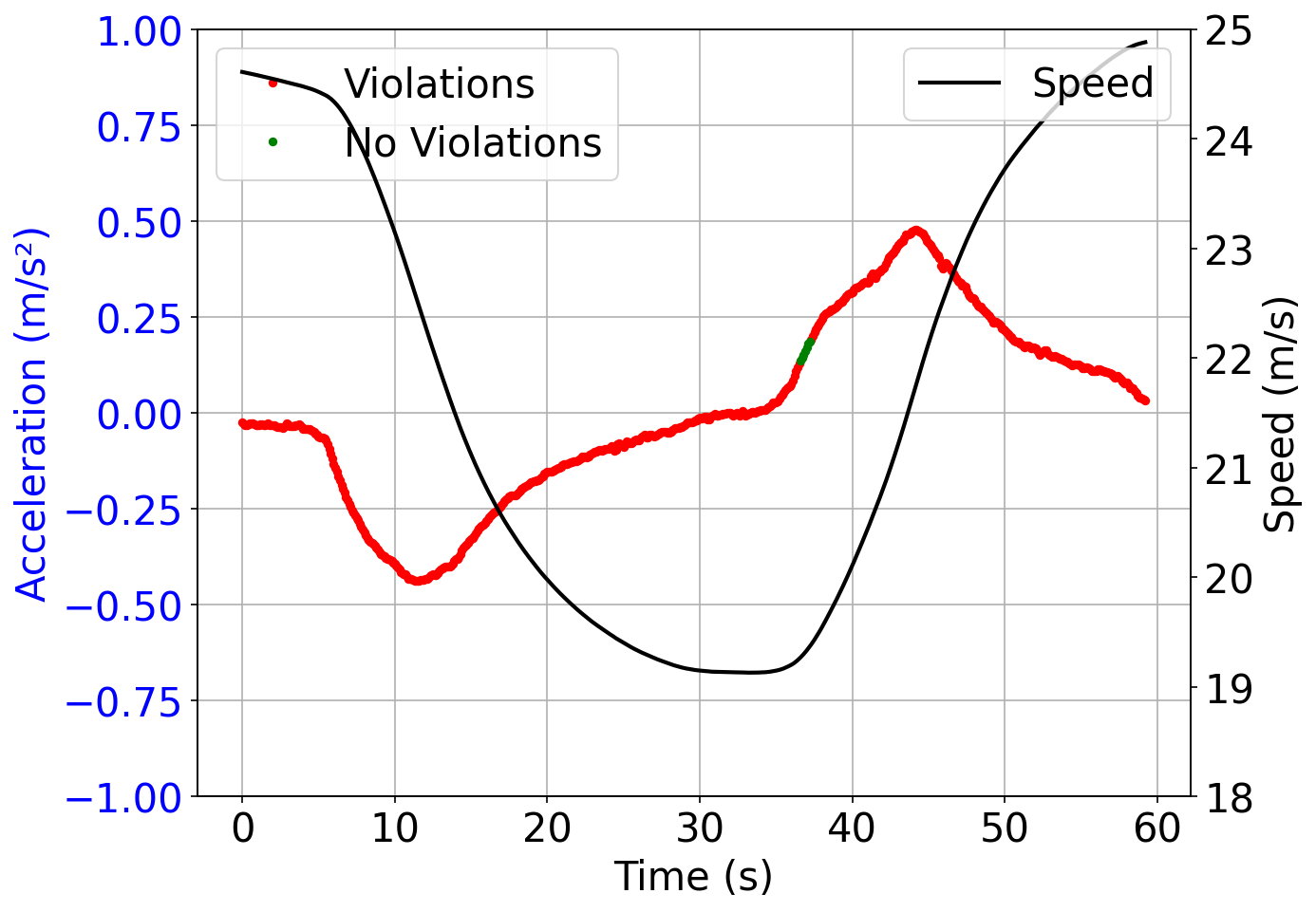}\label{rdc_2_1_max}}
\subfloat[][RDC Violations - Spacing]{\includegraphics[scale = 0.2]{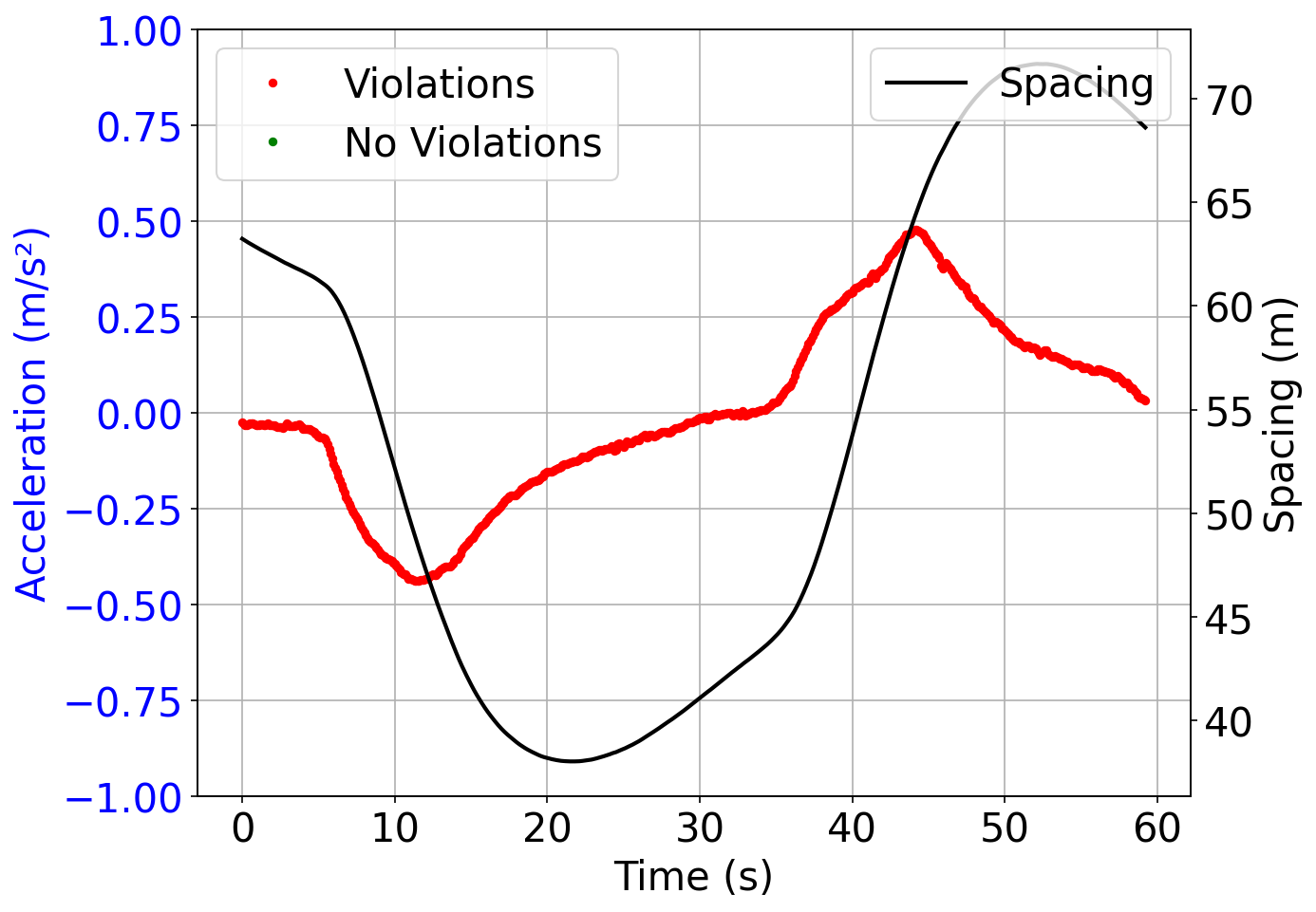}\label{rdc_2_2_max}}
\subfloat[][RDC Violations - Relative Speed]{\includegraphics[scale = 0.2]{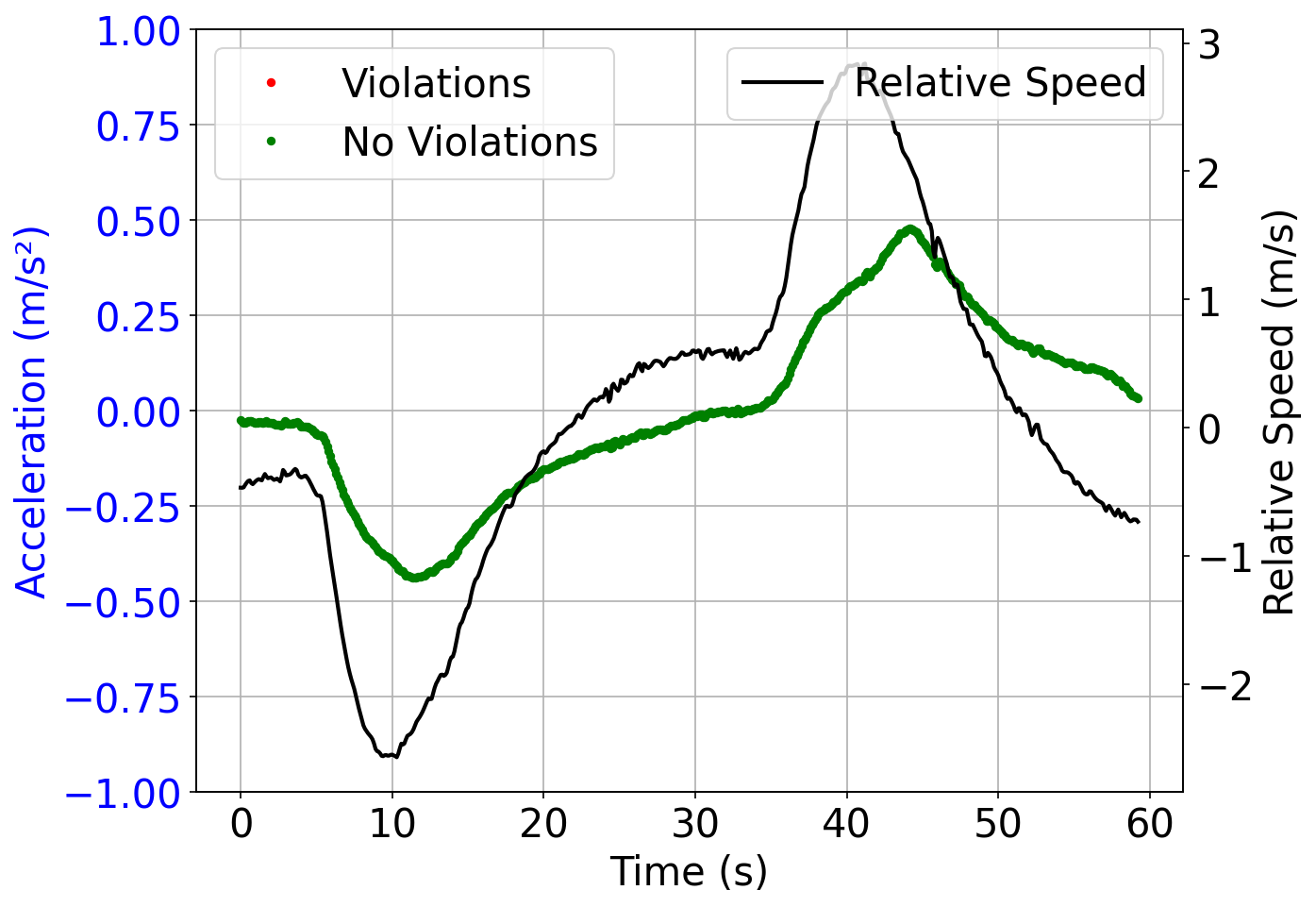}\label{rdc_2_3_max}}
\caption{Illustrations of predictions and violations by the physics-informed Neural Network model in terms of speed, spacing, and relative speed. The green and red dots denote predictions conforming to and violating the established rules, respectively.}
\label{fig:RDC_PINN_max_max}
\end{figure*}

\begin{figure*}
\centering
\subfloat[][RDCs Violations - Speed]{\includegraphics[scale = 0.2]{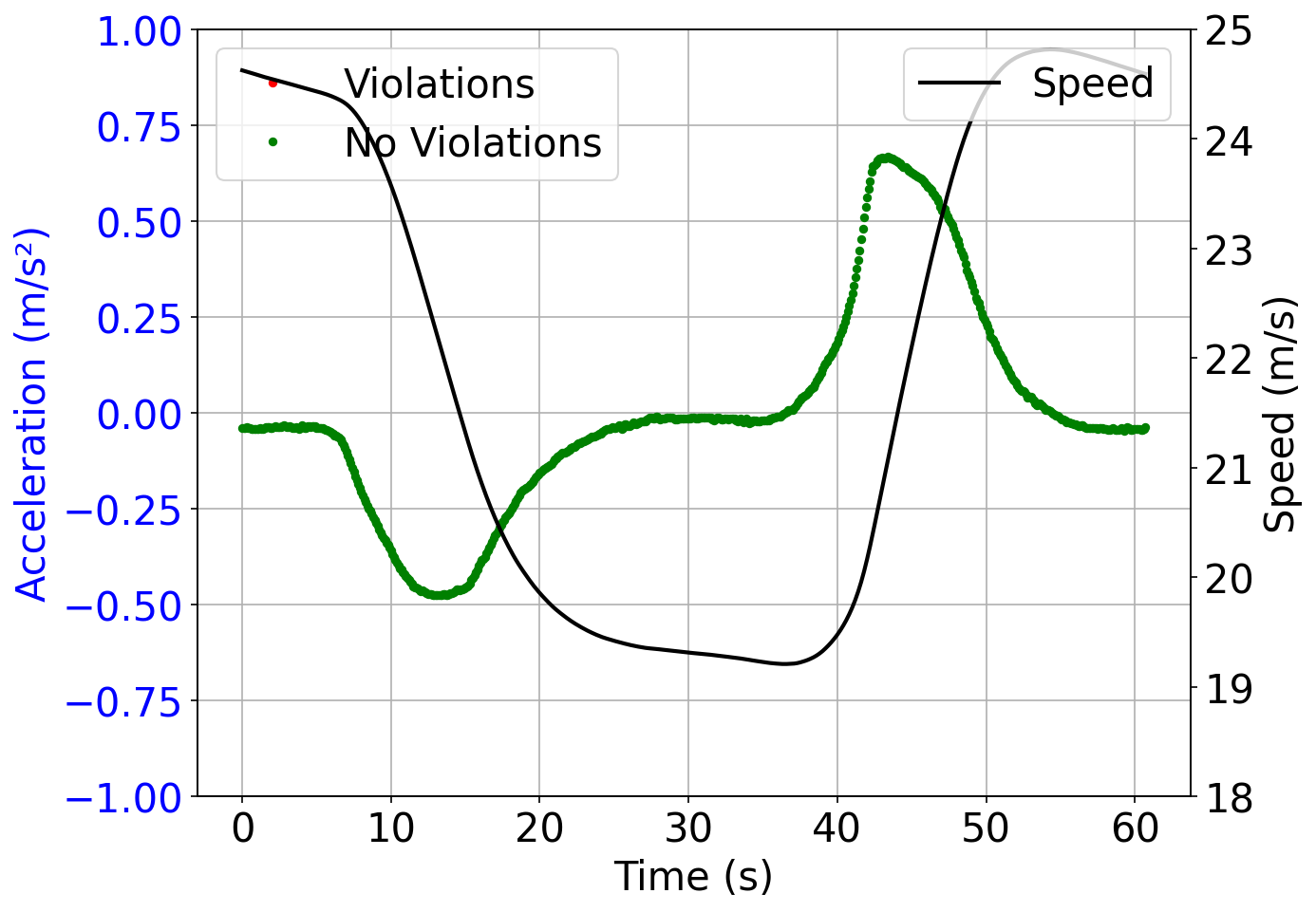}\label{rdc_3_1_max}}
\subfloat[][RDCs Violations - Spacing]{\includegraphics[scale = 0.2]{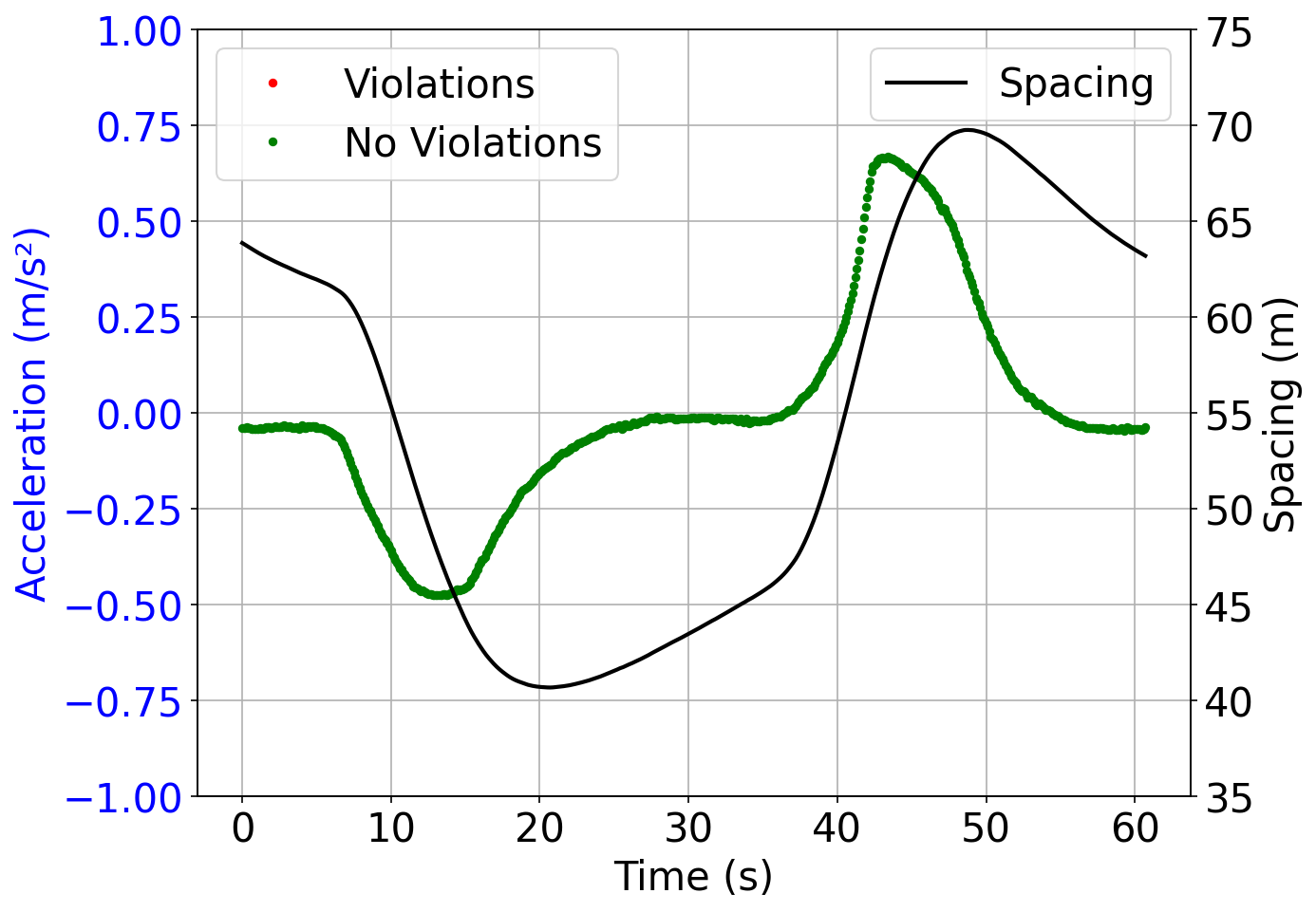}\label{rdc_3_2_max}}
\subfloat[][RDCs Violations - Relative Speed]{\includegraphics[scale = 0.2]{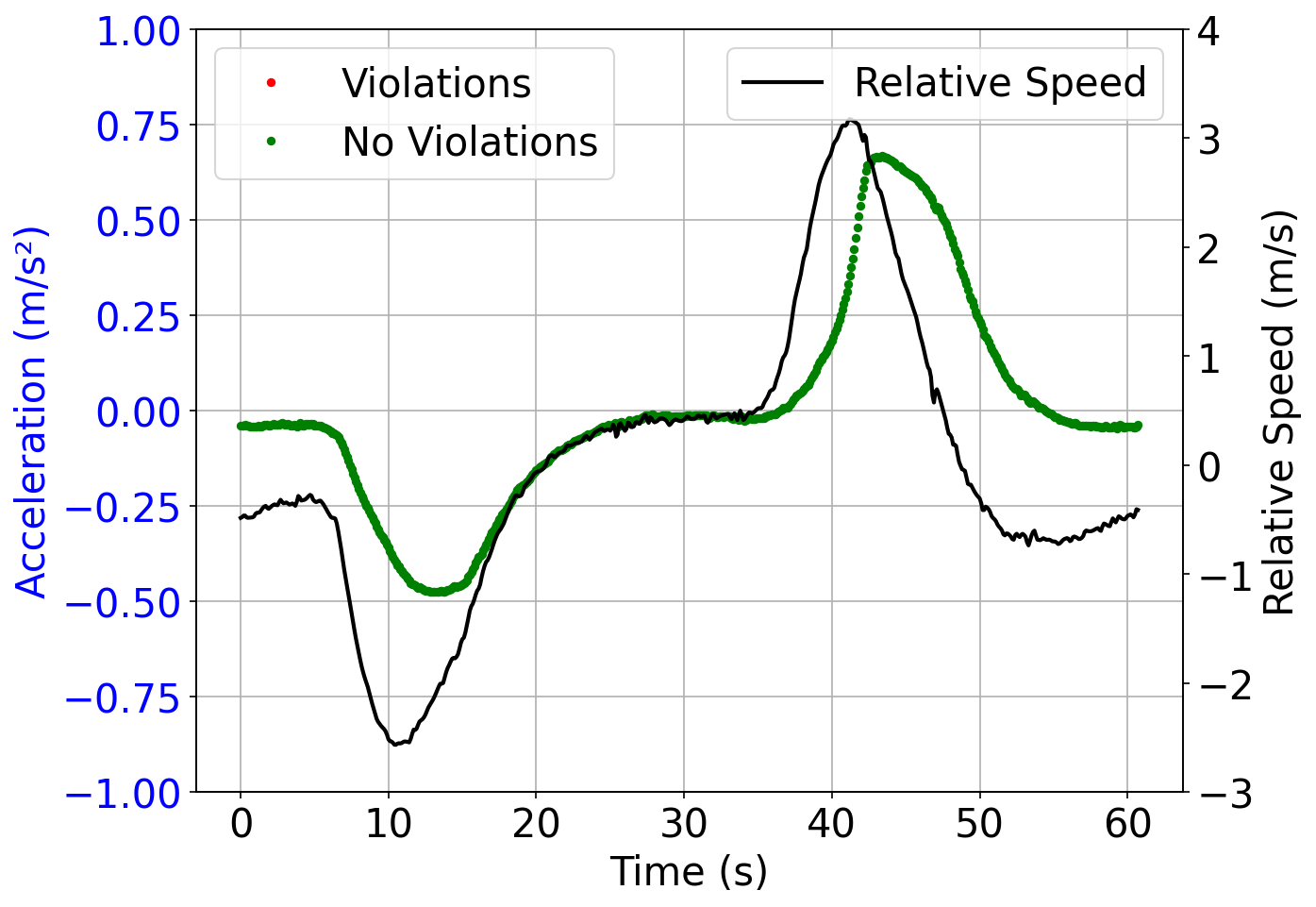}\label{rdc_3_3_max}}
\caption{Illustrations of predictions and violations by the RACER model in terms of speed, spacing, and relative speed. The green and red dots denote predictions conforming to and violating the established rules, respectively.}
\label{fig:RDC_NN_rational_max}
\end{figure*}

Fig.~\ref{fig:RDC_NN_max} to Fig.~\ref{fig:RDC_NN_rational_max} showcase the RDC violations for the ACC vehicle with the maximum setting. Consistent with the earlier results from the minimum setting comparison, the proposed RACER model is the only model that fully complies with the RDCs. In contrast, both the baseline NN and PINN models exhibit violations in one or more of the constraints. Specifically, the NN model fails in this scenario, leading to crashes and irrational predictions. These results further emphasize the advantage of incorporating RDCs into the learning process, as demonstrated by the superior performance of the RACER model.

\subsection{Model Evaluation: ACC Vehicle with Smoother Acceleration}

\begin{table}[ht]
\centering
\caption{Root mean squared errors for different models used as car-following controllers with smoother acceleration.}
    \begin{tabular}{lcccc}
        \toprule
        & \textbf{RACER} & \textbf{OVRV} & \textbf{NN} & \textbf{PINN} \\
        \midrule
    \textbf{Acceleration (m/s$^2$)} & \bf 0.099 & 0.111 & 0.115 & 0.111 \\
    \textbf{Speed (m/s)} & \bf 0.152 & 0.173 & 0.237 & 0.322 \\
    \textbf{Spacing (m)} & \bf 0.298 & 1.485 & 0.559 & 0.415 \\
    \bottomrule
\end{tabular}
\label{tab:rmse_results_smo}
\end{table}

In prior experiments, the actual vehicle acceleration was calculated using 0.1-second intervals, which introduced significant noise, as indicated by the high variance in the measurements. This noise obscured the assessment of our proposed RACER model's response to acceleration oscillations, challenging the verification of the model's rationality. To address this, we adopted a larger time interval of 0.5 seconds for a smoother estimation of the vehicle's average acceleration over that that longer time step with the minimum gap setting. Fig.~\ref{fig:RACER_smo} below illustrates the performance of various models. Notably, between 20 and 25 seconds, the RACER model uniquely adapts to the fluctuations in acceleration, influenced by the RDCs active during this period. Overall, the RACER model consistently surpasses competing models across all evaluated metrics, including acceleration, spacing, and speed as presented in Table~\ref{tab:rmse_results_smo}.

\begin{figure*}
\centering
\subfloat[][Model Performance - Acceleration]{\includegraphics[scale = 0.2]{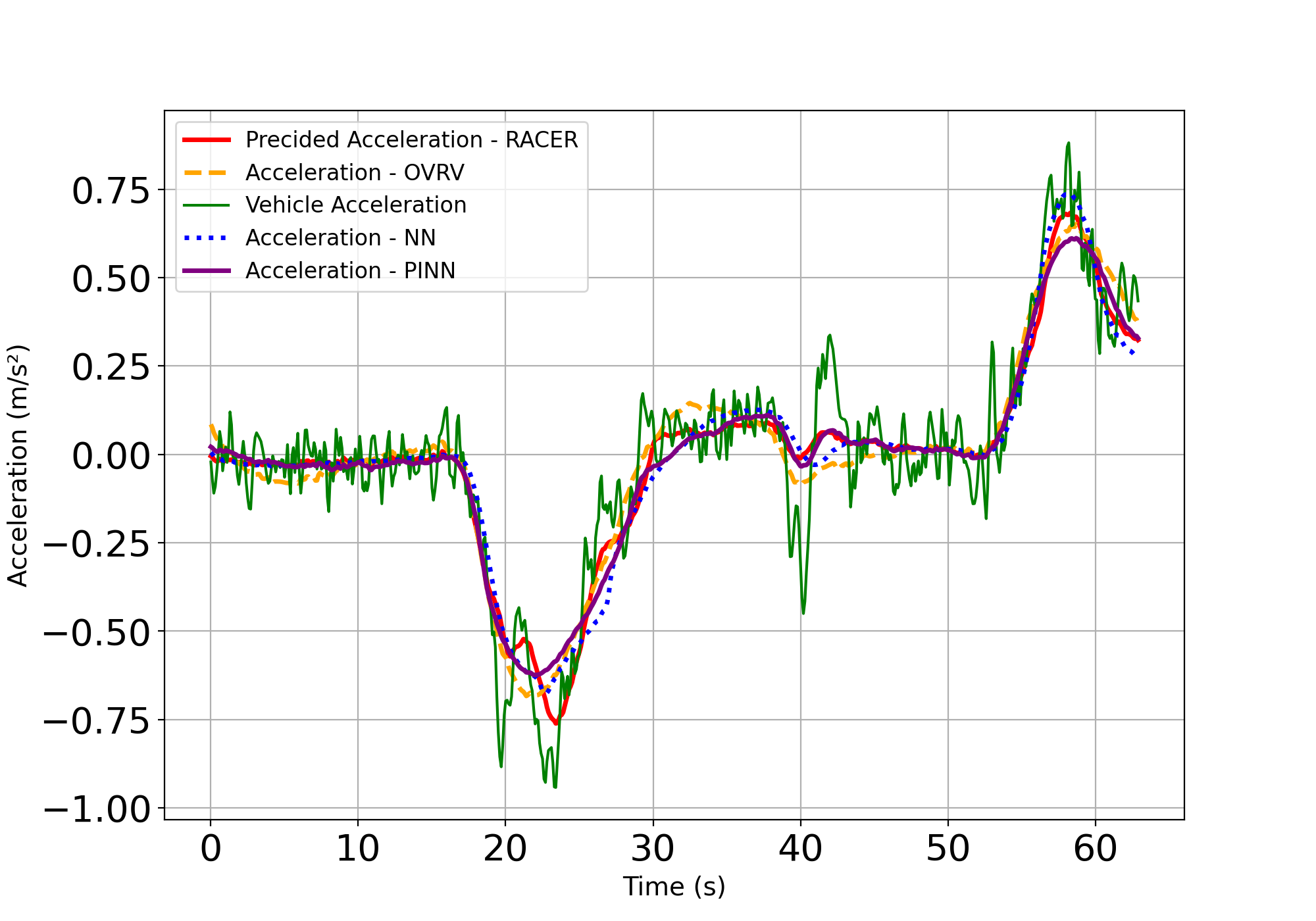}\label{fig:accel_compare_smo}}
\subfloat[][Model Performance - Spacing (controller)]{\includegraphics[scale = 0.2]{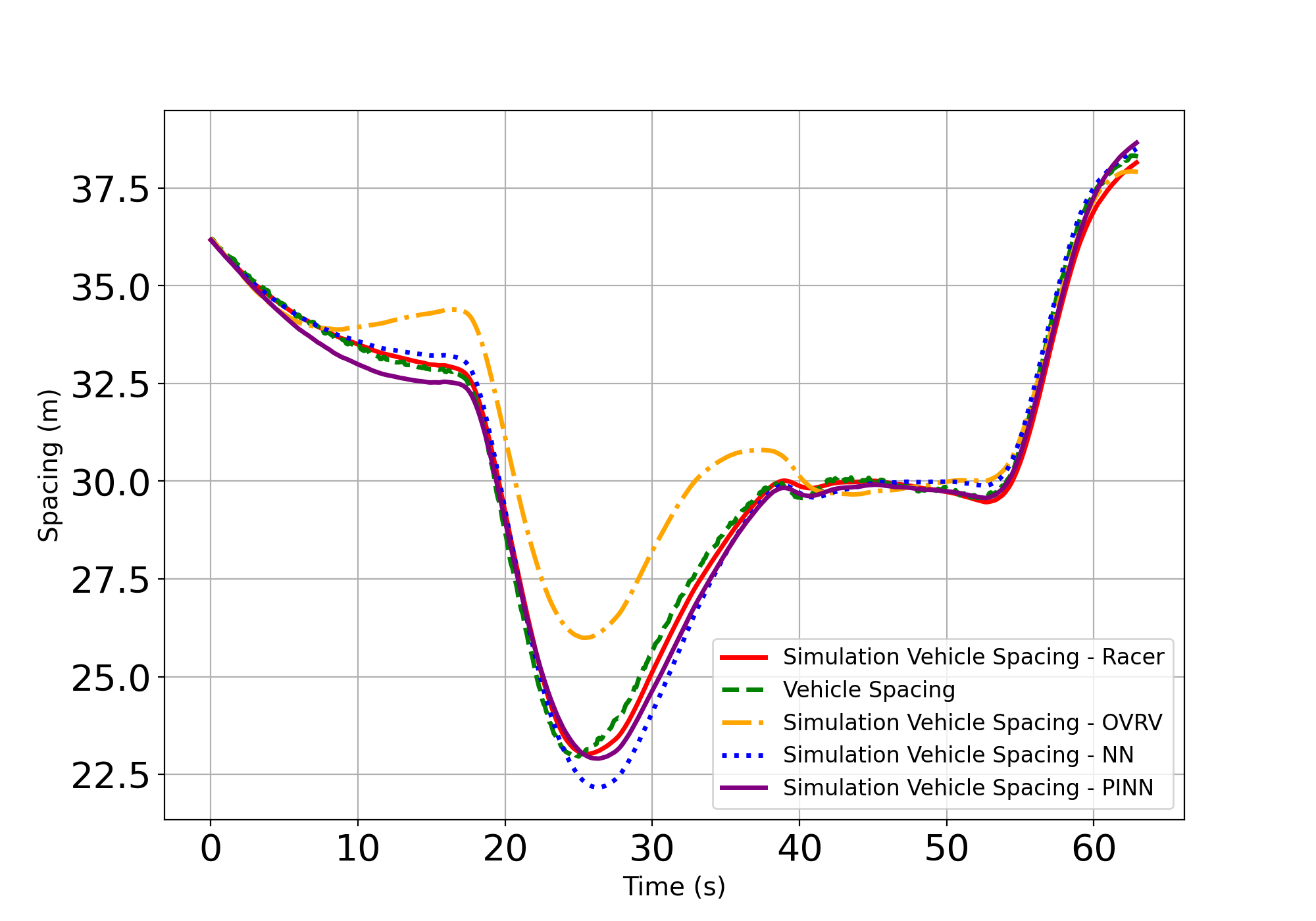}\label{fig:space_compare_smo}}
\subfloat[][Model Performance - Speed (controller)] {\includegraphics[scale = 0.2]{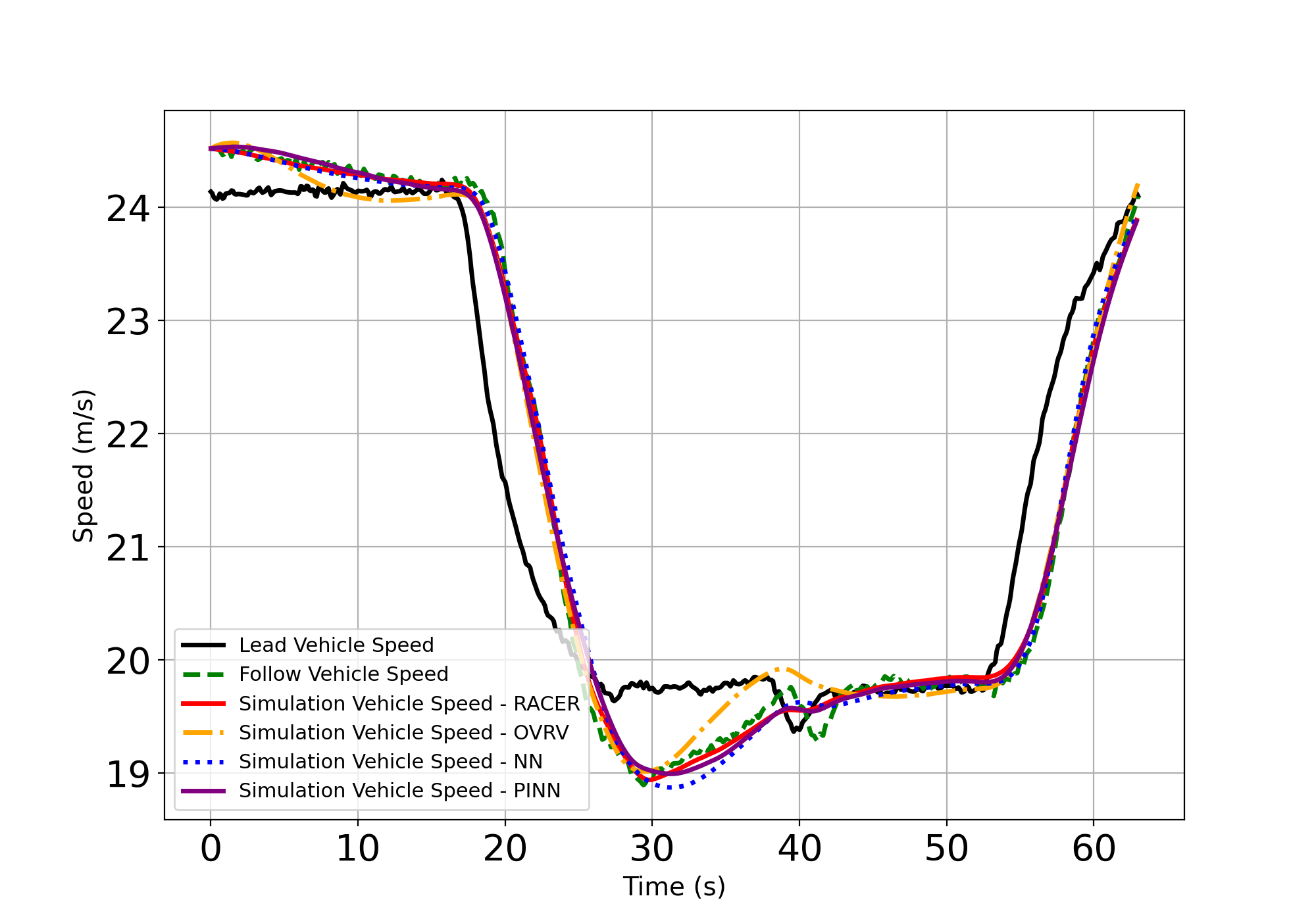}\label{fig:speed_compare_smo}}
\caption{Comparison of model performance with respect to acceleration, spacing, and speed.}
\label{fig:RACER_smo}
\end{figure*}

\begin{figure}
\centering
\includegraphics[scale=0.3]{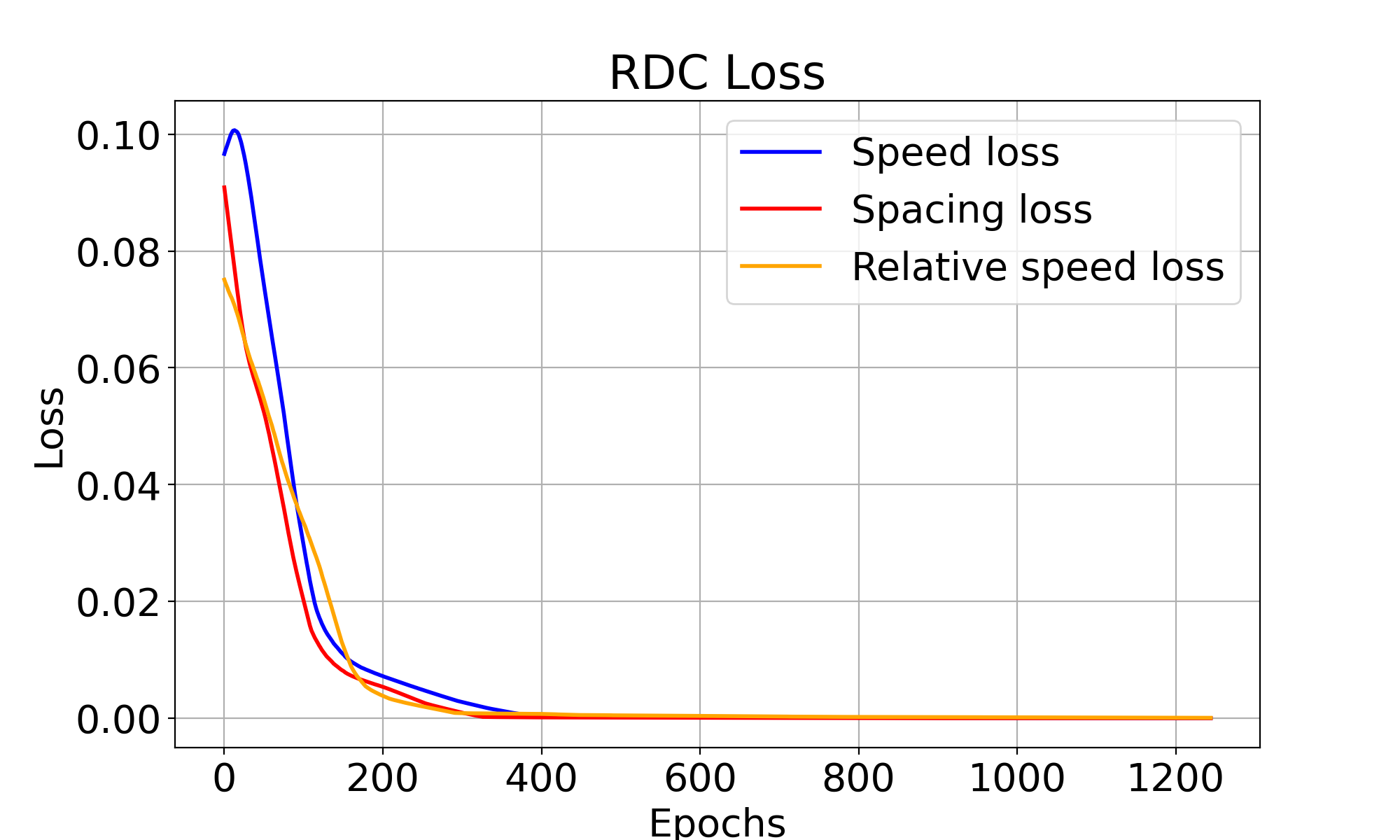}
\caption{Training progression of RDCs loss, detailing speed, spacing, and relative speed loss components.}
\label{fig:RDC_loss}
\end{figure}

In addition to the performance metrics discussed, the training process of the RACER model incorporates the minimization of the RDCs loss, which is crucial for maintaining safe and stable car-following behavior. Fig.~\ref{fig:RDC_loss} presents the RDCs loss progression throughout the training epochs. The loss is segmented into three components: speed loss, spacing loss, and relative speed loss. These components are critical as they correspond to the key aspects of rational car-following dynamics that our model aims to optimize.

During training, an evident and rapid decrease in RDC loss is observed, showcasing the model's ability to learn and adapt to the desired behavior efficiently. The convergence of the loss values indicates the model's proficiency in capturing the intrinsic patterns of vehicle dynamics for adaptive cruise control. It is particularly noteworthy that after an initial steep descent, the loss curves flatten out, suggesting that the model reaches a state of minimal error, which aligns with the desired outcome of the RACER model's learning process.

As Table \ref{tab:cf_comparison_alt} shows, the RACER model uniquely combines the predictive power of deep learning with the hard safety bounds imposed by Rational Driving Constraints. Unlike the OVRV model, RACER captures complex driving patterns; unlike the NN model, it ensures physically meaningful outputs; and unlike the PINN model, it achieves strict RDC compliance rather than relying on a soft physics residual. This combination of high accuracy and provable driving rationality is particularly valuable for ACC vehicles and other safety-critical applications, where even occasional irrational predictions could have hazardous consequences. The present study therefore represents an initial step toward seamlessly incorporating domain-specific knowledge into AI-based car-following models.

\begin{table*}[htbp]
\centering
\caption{Comparison of Car-Following Modeling Approaches.}
\label{tab:cf_comparison_alt}
\footnotesize
\begin{tabular}{@{}>{\centering\arraybackslash}p{2.5cm} >{\centering\arraybackslash}p{2.8cm} >{\centering\arraybackslash}p{2.8cm} >{\centering\arraybackslash}p{2.8cm} >{\centering\arraybackslash}p{2.8cm}@{}}
\toprule
\textbf{Characteristic} & \textbf{Physics-Based Models} & \textbf{Neural Networks} & \textbf{PINN} & \textbf{RACER} \\
\midrule
\textbf{Foundation} & 
Mathematical equations with physical principles & 
Data-driven learning from trajectories & 
Hybrid approach combining NN with physics models & 
NN with rational driving constraints \\
\addlinespace
\textbf{Physics Integration} & 
Hard-coded in model structure & 
None (learned implicitly) & 
Physics loss term in training & 
Derivative constraints in loss \\
\addlinespace
\textbf{RDC Compliance} & 
Directly calculable through PDEs of explicit equations & 
Not guaranteed & 
Not directly enforced & 
Explicitly enforced \\
\addlinespace
\textbf{RDC Performance} & 
Theoretically verifiable but limited model flexibility & 
High accuracy but potential violations & 
Good accuracy with some violations & 
High accuracy with zero violations \\
\addlinespace
\textbf{Safety Guarantees} & 
Theoretical through model structure & 
None & 
Partial through physics guidance & 
Complete through RDC enforcement \\
\bottomrule
\end{tabular}
\end{table*}

\section{Discussion \& Conclusion}\label{sec:con}

Our experimental findings highlight the significant differences between pure neural network models and their physically constrained counterparts, particularly with respect to handling nonlinear problems and offering predictions that surpass those provided by conventional models like OVRV. These results emphasize the critical importance of integrating physical constraints into neural networks used for acceleration predictions to ensure realistic long-term forecasts. In contrast to existing approaches, our hybrid model bridges the gap between engineering and data-driven methods, combining the reliability of physical constraints with the adaptability of machine learning. This hybrid approach enables the model to maintain the integrity of real-world driving principles while leveraging data to optimize its predictions. While recent literature, including studies like~\cite{punzo2021calibration, he2022physics}), has emphasized the calibration of car-following models based on spacing, our approach introduces a novel perspective by focusing on acceleration as the control output. This decision is grounded in real-world applications, where controllers rely on acceleration or velocity rather than spacing to dictate vehicle behavior. As such, our model's calibration aligns with the control dynamics observed in real-world ACC-equipped vehicles, ensuring a more practical and reliable implementation.

While the PINN framework has proven effective in implicitly capturing some of the outcomes of RDCs, our approach offers several key advancements by explicitly incorporating RDCs into the model's loss function. This explicit integration ensures that the model adheres to these constraints throughout the learning process, resulting in more reliable and interpretable predictions. The explicit incorporation of RDCs, as demonstrated in our RACER model, offers enhanced flexibility by allowing the model to learn complex patterns from real-world driving data while maintaining adherence to fundamental driving constraints. This capability is essential for accurately capturing the nuanced behaviors observed in real-world driving scenarios—behaviors that purely physics-based models may struggle to represent due to their simplified mathematical formulations. The zero RDC violations achieved by RACER result from the hard constraint enforcement mechanism where gradient-based penalties immediately correct any tendency toward irrational behavior during training. In contrast, other models exhibit systematic violation patterns: NN models violate spacing constraints when trying to minimize prediction error in dense traffic scenarios, while PINN models trade off between data fitting and physics compliance, leading to violations when these objectives conflict.

Implementing a loss function that embodies a physical model such as OVRV in a neural network may inadvertently impair performance by exacerbating divergent behaviors, especially when compared to relying on derivative constraints. Yet, with the judicious incorporation of RDCs into our model, we observed superior accuracy and performance, ultimately outshining other existing models. The trajectories generated by our model depict a more rational and safer driving behavior than those derived from alternative models, thereby boosting safety measures. This suggests that our proposed model can offer a novel paradigm for ACC driving behavior.

Incorporating physical constraints into neural network models has value, especially in accurately simulating the physical world. Our proposed controller aims to ensure the rationality of driving behavior. However, we do not assume that human driving behavior is always rational. This is why we focused our testing on ACC-equipped vehicles, which are automated vehicles that we assume should adhere to rational driving principles. The calibration of the RACER model relies on data from controlled experiments (Gunter et al.~\cite{gunter2020are}), where the leading vehicle's trajectory was pre-planned to elicit specific car-following patterns. While this approach enabled precise evaluation of the model under controlled conditions, it does not fully reflect the complexity of naturalistic driving scenarios. In real-world settings, ACC systems must respond to a variety of unpredictable behaviors, such as sudden braking when approaching a queue, lane changes by surrounding vehicles, or gaps created by diverging vehicles. These scenario-dependent factors may affect the ACC's performance in ways not accounted for in the current dataset. Consequently, while the model performs well within the scope of the tested conditions, its generalizability to more diverse real-world scenarios remains to be fully assessed. Therefore, the RMSE values reported in the Results Section should be regarded as best-case performance under controlled conditions.

Human driving behavior is undoubtedly more complex and drivers occasionally (and often briefly) deviate from the RDCs. However, our proposed model, which effectively combines physical-guided AI with car-following principles, exhibits potential to inform and guide human drivers, thus making driving safer and more rational. The model also holds promise for testing across a wide range of ACC vehicle driving scenarios as well as human driving situations. To enhance the robustness and applicability of the RACER model, future research should prioritize the collection and integration of naturalistic ACC car-following data. Such data would expose the model to a wider spectrum of driving behaviors, enabling a more comprehensive evaluation and refinement of its performance across varied real-world conditions. This includes scenarios involving unpredictable lead vehicle behavior, lane changes, and other complex traffic interactions that are essential for full real-world deployment. Future research could explore this exciting avenue and contribute to the progressive journey toward achieving safer, smarter, and more efficient transportation systems.

\section*{Author Contributions}
The authors confirm their contribution to the paper as follows: study conception and design: T. Li and R. Stern; data collection: T. Li; analysis and interpretation of results: T. Li, A. Halatsis, and R. Stern; draft manuscript preparation: T. Li, A. Halatsis, and R. Stern. All authors reviewed the results and approved the final version of the manuscript. The authors do not have any conflicts of interest to declare.

\section*{Acknowledgment}
This work is supported by the University of Minnesota Center for Transportation Studies through the Transportation Scholar's Program. T. Li acknowledges the support of the Dwight David Eisenhower Graduate Fellowship from the Federal Highway Administration. 

\section*{Declaration of Conflicting Interests}
The author(s) declared no potential conflicts of interest with respect to the research, authorship, and/or publication of this article.

\bibliographystyle{unsrt}  
\bibliography{refs}

\end{document}